\documentclass[10pt,twocolumn,letterpaper]{article}

\makeatletter
\@namedef{ver@everyshi.sty}{}
\makeatother
\usepackage[pagenumbers]{cvpr} 
\usepackage[accsupp]{axessibility}
\usepackage{graphicx}
\usepackage{amsmath}
\usepackage{amssymb}
\usepackage{booktabs}

\usepackage[pagebackref,breaklinks,colorlinks]{hyperref}

\usepackage[capitalize]{cleveref}
\crefname{section}{Sec.}{Secs.}
\Crefname{section}{Section}{Sections}
\Crefname{table}{Table}{Tables}
\crefname{table}{Tab.}{Tabs.}

\def\cvprPaperID{1024} 
\def\confName{CVPR}
\def\confYear{2023}

\usepackage{color}
\definecolor{url_color}{RGB}{42, 83, 163}
\usepackage{wrapfig}
\usepackage{setspace}
\usepackage{nicematrix}
\usepackage{array, multirow}
\usepackage{floatrow}
\usepackage{adjustbox}
\usepackage{pifont}
\usepackage{microtype}
\usepackage{mathptmx} 
\usepackage{times}

\DeclareMathOperator{\Exp}{Exp}

\DeclareMathOperator{\diag}{diag}
\newcommand{\norm}[1]{\left\lVert#1\right\rVert}
\newcommand\blfootnote[1]{%
  \begingroup
  \renewcommand\thefootnote{}\footnote{#1}%
  \addtocounter{footnote}{-1}%
  \endgroup
}

\begin{document}

\title{PVO: Panoptic Visual Odometry}

\author{
  Weicai Ye$^{1,2}$\thanks{Both authors contributed equally to this research.}\and
  Xinyue Lan$^{1,2}$\footnotemark[1]\and 
  Shuo Chen$^{1,2}$\and
  Yuhang Ming$^{3,4}$ \and 
  Xingyuan Yu$^{1,2}$\and
  Hujun Bao$^{1,2}$\and
  Zhaopeng Cui$^1$\and
  Guofeng Zhang$^{1,2}$\thanks{Corresponding author.}\and
  \textnormal{$^1$State Key Lab of CAD\&CG, Zhejiang University \quad $^2$ZJU-SenseTime Joint Lab of 3D Vision} 
  \\
 \textnormal{$^3$School of Computer Science, Hangzhou Dianzi University \quad $^4$VIL, University of Bristol} 
  \\
  \texttt{\small \{weicaiye, xinyuelan, chenshuo.eric, RickyYXY, baohujun, zhpcui, zhangguofeng\}@zju.edu.cn}\\
  \texttt{\small  yuhang.ming@hdu.edu.cn}  
  \\
  \texttt{\small \urlstyle{tt}\textcolor{url_color}{\url{https://zju3dv.github.io/pvo/}}}
}

\twocolumn[{%
    \renewcommand\twocolumn[1][]{#1}%
    \setlength{\tabcolsep}{0.0mm} 
    \maketitle
    \begin{center}
        \newcommand{\teaserwidth}{\textwidth}
    \vspace{-0.4in}
        \includegraphics[width=\linewidth]{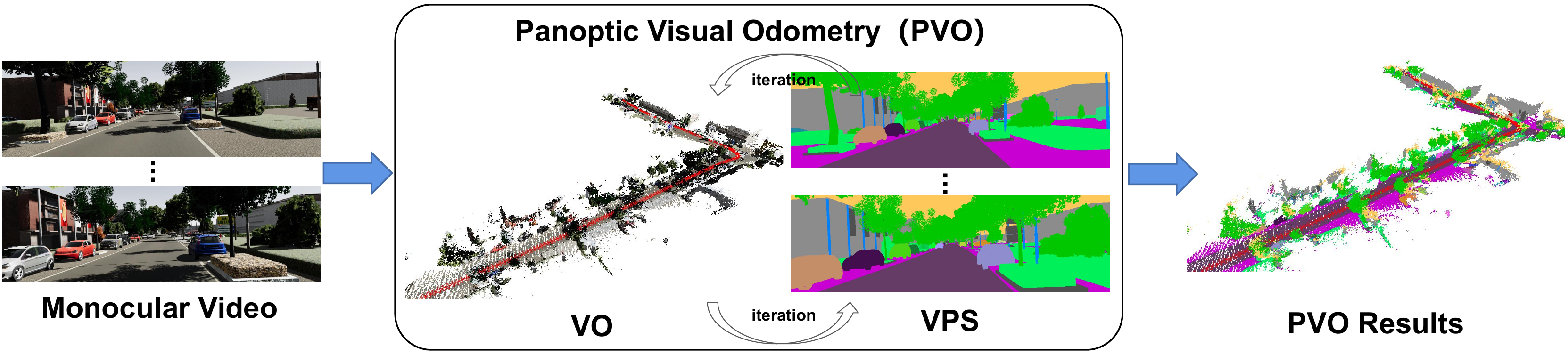}
      \vspace{-0.2in}
        \captionof{figure}{\textbf{Panoptic Visual Odometry.} PVO takes monocular video as input and outputs the panoptic 3D map while simultaneously localizing the camera itself with respect to the map.}
    \label{fig:teaser}
    \vspace{-0.2em}
    \end{center}
}]

\maketitle
\begin{abstract}
\vspace{-0.6em}
We present PVO, a novel panoptic visual odometry framework to achieve more comprehensive modeling of the scene motion, geometry, and panoptic segmentation information. Our PVO models visual odometry (VO) and video panoptic segmentation (VPS) in a unified view, which makes the two tasks mutually beneficial. Specifically, we introduce a panoptic update module into the VO Module with the guidance of image panoptic segmentation. This Panoptic-Enhanced VO Module can alleviate the impact of dynamic objects in the camera pose estimation with a panoptic-aware dynamic mask. On the other hand, the VO-Enhanced VPS Module also improves the segmentation accuracy by fusing the panoptic segmentation result of the current frame on the fly to the adjacent frames, using geometric information such as camera pose, depth, and optical flow obtained from the VO Module. These two modules contribute to each other through recurrent iterative optimization. Extensive experiments demonstrate that PVO outperforms state-of-the-art methods in both visual odometry and video panoptic segmentation tasks.
   
\end{abstract}
\blfootnote{$\ast$ indicates equal contribution. $^\dagger$ indicates the corresponding author.}

\vspace{-1em}
\section{Introduction}
\label{sec:intro}
Understanding the motion, geometry, and panoptic segmentation of the scene plays a crucial role in computer vision and robotics, with applications ranging from autonomous driving to augmented reality. In this work, we take a step toward solving this problem to achieve a more comprehensive modeling of the scene with monocular videos.

\begin{figure}
  \centering
  \includegraphics[width=\linewidth]{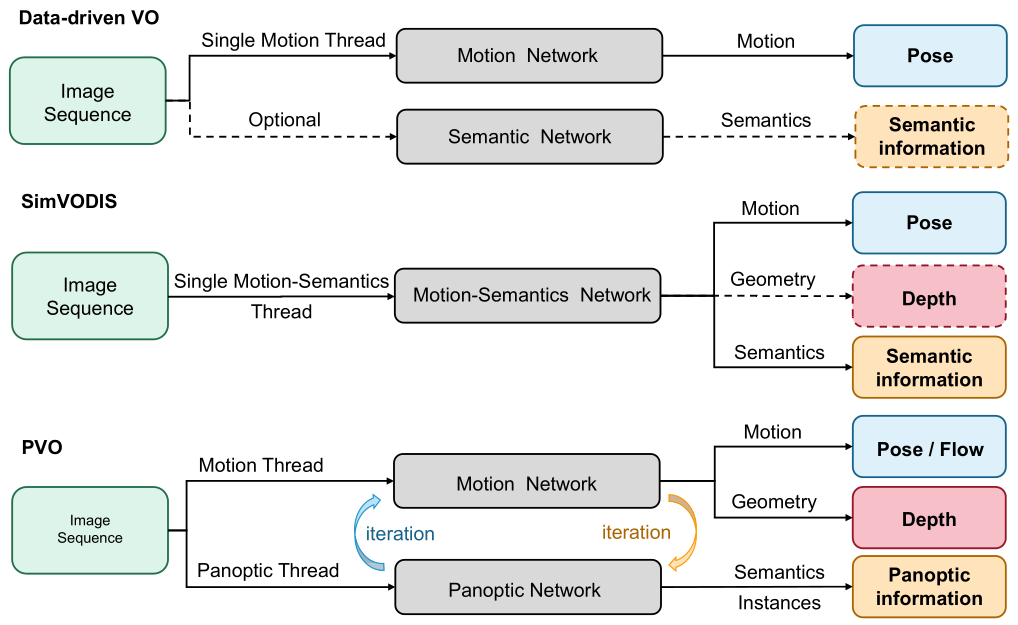}
  \caption{\textbf{Illustration.} Our PVO unifies visual odometry and video panoptic segmentation so that the two tasks can be mutually reinforced by iterative optimization. In contrast, methods such as SimVODIS~\cite{kim2019simvodis} optimize motion and semantic information in a multi-task manner.}
  \label{fig: illustration}
\end{figure}

Two tasks have been proposed to address this problem, namely visual odometry (VO) and video panoptic segmentation (VPS).
In particular, VO~\cite{engel2013semi, forster2014svo, deepvo} takes monocular videos as input and estimates the camera poses under the static scene assumption. To handle dynamic objects in the scene, some dynamic SLAM systems~\cite{bescos2018dynaslam, xiao2019dynamic} use instance segmentation network~\cite{He_2017_ICCV} for segmentation and explicitly filter out certain classes of objects, which are potentially dynamic, such as pedestrians or vehicles.
However, such approaches ignore the fact that potentially dynamic objects can actually be stationary in the scene, such as a parked vehicle.
In contrast, VPS~\cite{kim2020video, woo2021learning, Ye2022Hybrid} focuses on tracking individual instances in the scene across video frames given some initial panoptic segmentation results. Current VPS methods do not explicitly distinguish whether the object instance is moving or not. Although existing approaches broadly solve these two tasks independently, it is worth noticing that dynamic objects in the scene can make both tasks challenging.
Recognizing this relevance between the two tasks, some methods~\cite{ding2020every, cheng2017segflow, li2021unsupervised, kim2019simvodis} try to tackle both tasks simultaneously and train motion-semantics networks in a multi-task manner, shown in Fig.~\ref{fig: illustration}. However, the loss functions used in these approaches may contradict each other, thus leading to performance drops.

In this work, we propose a novel panoptic visual odometry (PVO) framework that tightly couples these two tasks using a unified view to model the scene comprehensively.
Our insight is that VPS can adjust the weight of VO with panoptic segmentation information (the weights of the pixels of each instance should be correlated) and VO can convert the tracking and fusion of video panoptic segmentation from 2D to 3D. Inspired by the seminal Expectation-Maximization algorithm~\cite{EM1996}, recurrent iterative optimization strategy can make these two tasks mutually beneficial.

Our PVO consists of three modules, an image panoptic segmentation module, a Panoptic-Enhanced VO Module, and a VO-Enhanced VPS Module. Specifically, the panoptic segmentation module (see Sec.~\ref{sec:panoptic seg}) takes in single images and outputs the image panoptic segmentation results, which are then fed into the Panoptic-Enhanced VO Module as initialization. Note that although we choose PanopticFPN~\cite{kirillov2019panoptic}, any segmentation model can be used in the panoptic segmentation module. In the Panoptic-Enhanced VO Module (see Sec.~\ref{sec:vps-enhanced vo}), we propose a panoptic update module to filter out the interference of dynamic objects and hence improve the accuracy of pose estimation in the dynamic scene. In the VO-Enhanced VPS Module (see Sec.~\ref{sec:vo-enhanced vps}), we introduce an online fusion mechanism to align the multi-resolution features of the current frame to the adjacent frames based on the estimated pose, depth, and optical flow. This online fusion mechanism can effectively solve the problem of multiple object occlusion. Experiments show that the recurrent iterative optimization strategy improves the performance of both VO and VPS. Overall, our contributions are summarized as four-fold. 
\begin{itemize}
    \item We present a novel Panoptic Visual Odometry (PVO) framework, which can unify VO and VPS tasks to model the scene comprehensively.
    \item A panoptic update module is introduced and incorporated into the Panoptic-Enhanced VO Module to improve pose estimation.
    \item An online fusion mechanism is proposed in the VO-Enhanced VPS Module, which helps to improve video panoptic segmentation.
    \item Extensive experiments demonstrate that the proposed PVO with recurrent iterative optimization is superior to state-of-the-art methods in both visual odometry and video panoptic segmentation tasks.
\end{itemize}


\vspace{-1em}
\section{Related Work}
\subsection{Video Panoptic Segmentation}
Video panoptic segmentation aims to generate consistent panoptic segmentation and track the instances to all pixels across video frames. A pioneer work, VPSNet~\cite{kim2020video} defines this novel task and proposes an instance-level tracking-based approach. SiamTrack~\cite{woo2021learning} extends VPSNet by proposing a pixel-tube matching loss and a contrast loss to improve the discriminative power of instance embedding. VIP-Deeplab~\cite{qiao2021vip} presents a depth-aware VPS network by introducing additional depth information.
While STEP~\cite{weber2021step} proposes to segment and track every pixel for video panoptic segmentation. HybridTracker~\cite{Ye2022Hybrid} proposes to track instances from two perspectives: the feature space and the spatial location. Different from existing methods, we introduce a VO-Enhanced VPS Module, which exploits the camera pose, depth, and optical flow estimated from VO to track and fuse information from the current frame to the adjacent frames, and can handle occlusion. 
\begin{figure*}[t]
    \centering
    \includegraphics[width=\textwidth]{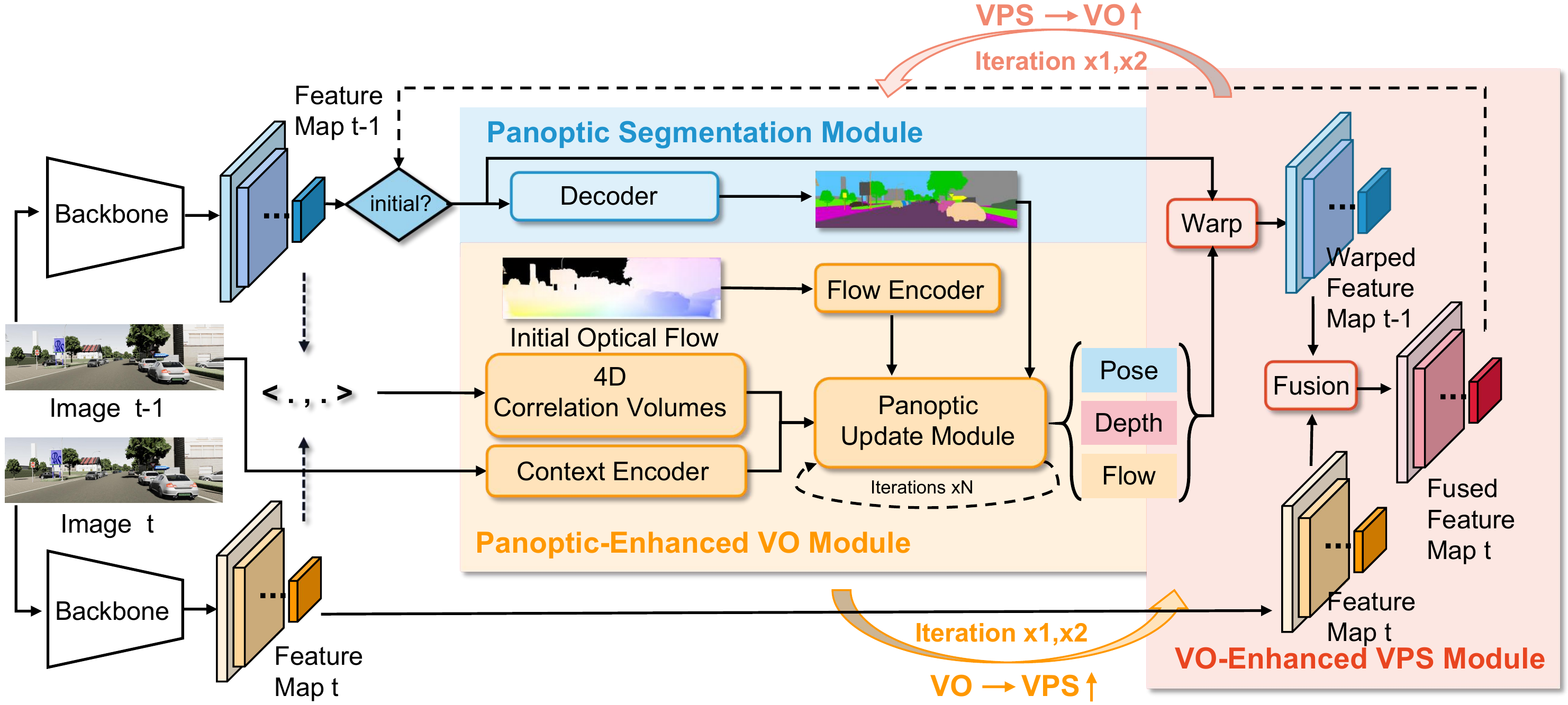}
    \caption{\textbf{Panoptic Visual Odometry Framework.} Our method consists of three modules, namely, an image panoptic segmentation module for system initialization (blue), a Panoptic-Enhanced VO Module (orange), and a VO-Enhanced VPS Module (red). The last two modules contribute to each other in a recurrent iterative manner.
    }
    \label{fig:pipeline}
    \vspace{-0.8em}
\end{figure*}

\subsection{SLAM and Visual Odometry}
SLAM stands for simultaneous self-localization and map construction, and visual odometry, serving as the front end of SLAM, focuses on pose estimation.
Modern SLAM systems roughly fall into two categories, geometry-based methods~\cite{Engel2014lsd, mur2015orb, forster2014svo, yang2022fdslam}, and learning-based methods~\cite{Zhou2017SfMLearner, Vija2017SfMNet, wang2018end, wang2019improving}. With the promising performance of supervised learning-based methods, unsupervised learning-based VO methods~\cite{yin2018geonet, zhan2018unsupervised, Ranjan_2019_CVPR} have received much attention, but they do not perform as well as supervised ones. Some unsupervised methods~\cite{Yang_2020_CVPR, Zou_2018_ECCV, jiao2021effiscene} exploit multi-task learning with auxiliary tasks such as depth and optical flow to improve performance.

Recently, TartanVO~\cite{tartanvo} proposes to build a generalizable learning-based VO and tests the system on a challenging SLAM dataset, TartanAir~\cite{tartanair2020iros}. DROID-SLAM~\cite{teed2021droid} proposes to iteratively update the camera pose and pixel-wise depth with a dense bundle adjustment layer and demonstrates superior performance. DeFlowSLAM~\cite{Ye2022DeFlowSLAM} further proposes dual-flow representation and a self-supervised method to improve the performance of the SLAM system in dynamic scenes. To tackle the challenge of dynamic scenes, dynamic SLAM systems~\cite{gonzalez2022s, chen2019suma++} usually leverage semantic information as constraints~\cite{lianos2018vso} or prior to improve the performance of the conventional geometric-based SLAM, but they~\cite{xu2019midfusion, runz2018maskfusion, fan2022blitz, ming2021objreloc, zhang2020vdo, bescos2021dynaslam, bescos2018dynaslam, zhu2022fusing, ou2022indoor} mostly act on the stereo, RGBD, or LiDAR sequences. 
Instead, we introduce a panoptic update module
and build the panoptic-enhanced VO on DROID-SLAM, and can work on monocular videos. Such a combination makes it possible to better understand of scene geometry and semantics, hence more robust to the dynamic objects in the scenes. Unlike other multi-task end-to-end models~\cite{kim2019simvodis}, our PVO has a recurrent iterative optimization strategy that prevents the tasks from jeopardizing each other. 



\vspace{-1em}
\section{Method}
Given a monocular video, PVO aims for simultaneous localization and panoptic 3D mapping.
Fig.~\ref{fig:pipeline} depicts the framework of the PVO model. It consists of three main modules: an image panoptic segmentation module, a Panoptic-Enhanced VO Module, and a VO-Enhanced VPS Module. The VO Module aims at estimating camera pose, depth, and optical flow, while the VPS Module outputs the corresponding video panoptic segmentation. The last two modules contribute to each other in a recurrent interactive manner.

\subsection{Image Panoptic Segmentation}
\label{sec:panoptic seg}
Image panoptic segmentation takes single images as input, and outputs the panoptic segmentation results of the images, which combines semantic segmentation and instance segmentation to model the instances of the image comprehensively. The output result is used to initialize video panoptic segmentation and then fed into the Panoptic-Enhanced VO Module (see Sec.~\ref{sec:vps-enhanced vo}). In our experiments, if not specifically indicated, we use the widely-used image panoptic segmentation network, PanopticFPN~\cite{kirillov2019panoptic}. PanopticFPN is built on the backbone of ResNet $f_{\theta_e}$ with weight $\theta_e$ and extracts multi-scale features of image $I_t$:
\begin{equation}
    \mathbf{z}_t = f_{\theta_e}(I_t)
    \label{eq:latent_code}
\end{equation}
It outputs the panoptic segmentation results using a decoder $g_{\theta_d}$ with weights $\theta_d$, consisting of semantic segmentation and instance segmentation. The panoptic segmentation results of each pixel $\mathbf{p}$ are: 
\begin{equation}
    P_{s}(\mathbf{p}|\mathbf{z}_t) = g_{\theta_d}(\mathbf{p}, \mathbf{z}_t)
    \label{eq:ps_prediction}
\end{equation}

The multi-scale features which are fed into the decoder are updated over time. In the beginning, the multi-scale features generated by the encoder are directly fed into the decoder (Fig.~\ref{fig:pipeline} blue part). In the later timesteps, these multi-scale features are updated with the online feature fusion module before being fed into the decoder (see Sec.~\ref{sec:vo-enhanced vps online_fusion}).

\subsection{Panoptic-Enhanced VO Module}
\label{sec:vps-enhanced vo}

\begin{figure}
  \centering
  \vspace{-1em}
  \includegraphics[width=\linewidth]{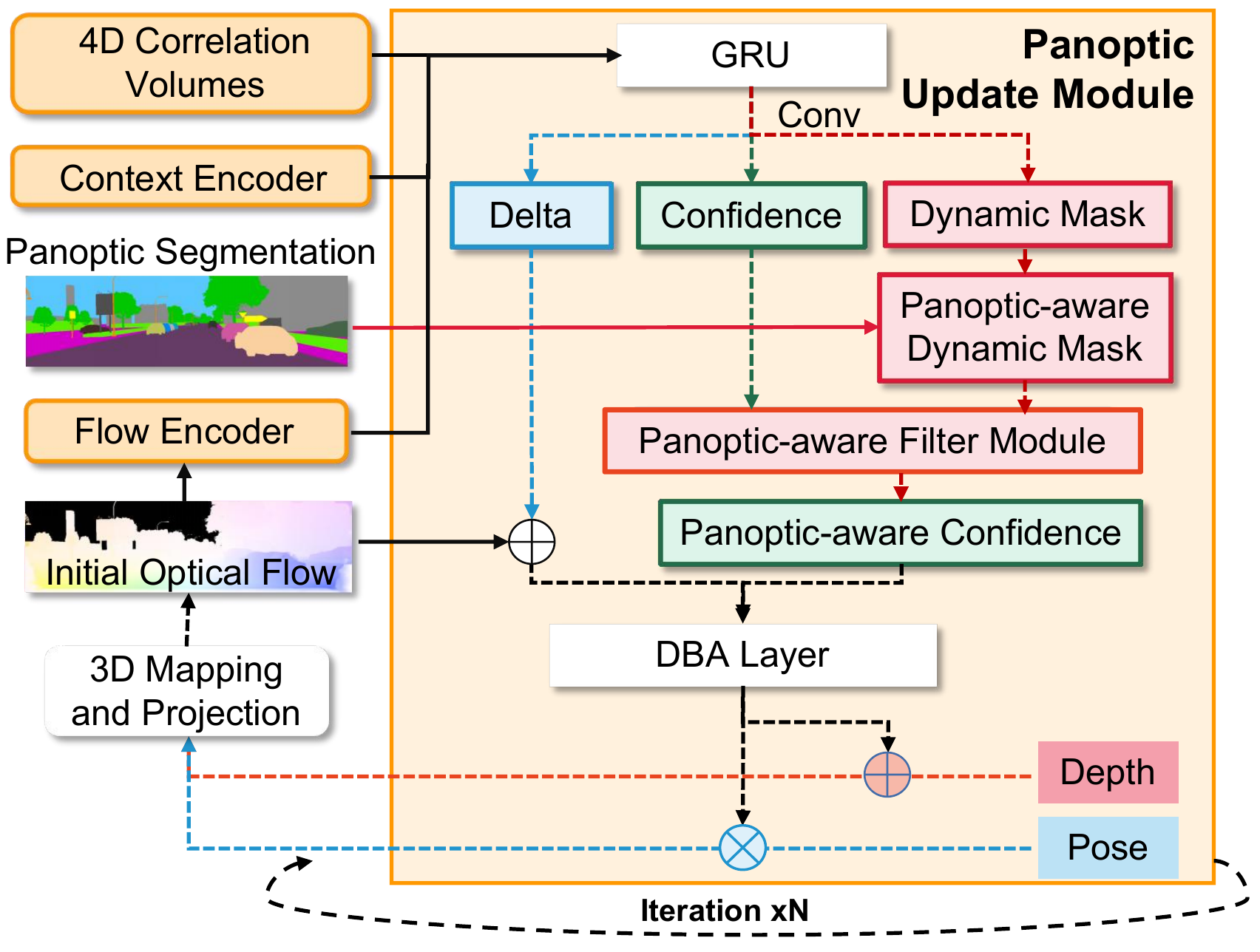}
  \caption{\textbf{Panoptic-Enhanced VO Module.} 
    The Panoptic-Enhanced VO Module mainly feeds the 4D correlation volumes, the context information from the context encoder, and the flow information into the panoptic update module. The panoptic update module iterates N times to obtain better depth, pose, and optical flow estimation. The panoptic segmentation information is used to adjust the correlation weight and the optical flow is initialized as 0 and iteratively updated with the DBA layer.} 
    \label{fig:Panoptic-enhanced_vo_module}
  \vspace{-1ex}
\end{figure}

In visual odometry, where dynamic scenes are ubiquitous, it is crucial to filter out the interference of dynamic objects. The front-end of DROID-SLAM~\cite{teed2021droid} takes monocular video $\{\mathbf{I}_t\}_{t=0}^N$ as input and optimizes the residuals of camera pose $\{\mathbf{G}_t\}_{t=0}^N \in SE(3)$ and inverse depth $\mathbf{d}_t \in \mathbb{R}_{+}^{H\times W}$ by iteratively optimizing optical flow delta $\mathbf{r}_{ij} \in \mathbb{R}^{H \times W \times 2}$ with confidence $\mathbf{w}_{ij} \in \mathbb{R}^{H \times W \times 2}$. 
It does not consider that most backgrounds are static, foreground objects may be dynamic, and the weights of the pixels of each object should be correlated. The insight of the Panoptic-Enhanced VO Module (see Fig.~\ref{fig:Panoptic-enhanced_vo_module}) is to assist in obtaining better confidence estimation (see Fig.~\ref{fig:Panoptic-Aware Weight}), by incorporating information from the panoptic segmentation. Thus, Panoptic-Enhanced VO can get more accurate camera poses. Next, we will briefly review the similar part (feature extraction and correlation) with DROID-SLAM, and focus on the sophisticated design of the panoptic update module.

\subsubsection{Feature Extraction and Correlation}
\textbf{Feature Extraction.} Similar to DROID-SLAM~\cite{teed2021droid}, the Panoptic-Enhanced VO Module borrows the key components of RAFT~\cite{raft} to extract the features. We use two separate networks (a feature encoder and a context encoder) to extract the multi-scale features of each image, where the features from the feature encoder are exploited to construct 4D correlation volumes of pair images, and the features from the context encoder are injected into the panoptic update module (see Sec.~\ref{sec: Panoptic Update Module}). The structure of the feature encoder is similar to the backbone of the panoptic segmentation network, and they can use a shared encoder. Note that for implementation convenience, we use different encoders.


\noindent \textbf{Correlation Pyramid and Lookup.} Similar to DROID-SLAM~\cite{teed2021droid}, we adopt a frame graph $(\mathcal{V}, \mathcal{E})$ to indicate the co-visibility between frames. 
For example, an edge $(i,j) \in \mathcal{E}$ represents the two images $I_i$ and $I_j$ maintaining overlapped areas, and a 4D correlation volume can be constructed through dot product between the feature vectors of these two images:
\begin{equation}
    C^{ij} = \langle g_\theta(I_i), g_\theta(I_j)\rangle
\end{equation}
The average pooling layer is followed to gain the pyramid correlation. We use the same lookup operator defined in DROID-SLAM~\cite{teed2021droid} to index the pyramid correlation volume values with bilinear interpolation. These correlation features are concatenated, resulting in the final feature vectors. 

\subsubsection{Panoptic Update Module}
\label{sec: Panoptic Update Module}
The Panoptic-Enhanced VO Module (see Fig.~\ref{fig:Panoptic-enhanced_vo_module}) which inherits from the front-end VO Module of DROID-SLAM, leverages the panoptic segmentation information to adjust the weight of VO. The flow information obtained by feeding the initial optical flow to the flow encoder
and the 4D correlation volumes established from the two frames and the features acquired by the context encoder are fed to the GRU as intermediate variables, and then the three convolutional layers output a dynamic mask $\mathbf{M_d}_{ij} \in \mathbb{R}^{H \times W \times 2}$, a correlation confidence map $\mathbf{w}_{ij} \in \mathbb{R}^{H \times W \times 2}$ and a dense optical flow delta $\mathbf{r}_{ij} \in \mathbb{R}^{H \times W \times 2}$, respectively. We can adjust the dynamic mask to the panoptic-aware dynamic mask given the initialized panoptic segmentation. For understanding, we leave the notation unchanged. Especially, the stuff segmentation will be set as static, while the foreground objects with high dynamic probability will be set as dynamic. The confidence and panoptic-aware dynamic mask are passed through a panoptic-aware filter module to obtain the panoptic-aware confidence: 
\begin{equation}
    \mathbf{w_p}_{ij}=\operatorname{sigmoid}(\mathbf{w}_{ij}+(1-\mathbf{M_d}_{ij})\cdot\eta)
\end{equation}
where $\eta$ is set as 10 in our experiment.

The obtained flow delta $\mathbf{r}_{ij}$ adding the original optical flow is fed to the dense bundle adjustment (DBA) layer to optimize the residual of the inverse depth and the pose.
The panoptic update module is iteratively optimized $N$ times until convergence. Following DROID-SLAM~\cite{teed2021droid}, the pose residuals $\Delta \boldsymbol \xi^{(n)}$ are transformed on the SE3 manifold to update the current pose, while the residuals of depth and dynamic mask are added to the current depth and dynamic mask, respectively:
\begin{equation}
     \mathbf{G}^{(n+1)} = \Exp(\Delta \boldsymbol \xi^{(n)}) \circ \mathbf{G}^{(n)}
\end{equation}
\begin{equation}
     \Theta^{(n+1)} = \Delta \Theta^{(n)} + \Theta^{(n)}, \Theta \in \{\mathbf{d},\mathbf{M_d}\}
\end{equation}

\noindent\textbf{Correspondence.} We first use the current pose and depth estimates at each iteration to search for the correspondence. Refer to DROID-SLAM~\cite{teed2021droid}, for each pixel coordinates $\mathbf{p}_i \in \mathbb{R}^{H \times W \times 2}$ in frame $i$, the dense correspondence field $\mathbf{p}_{ij}$ for each edge $(i, j) \in \mathcal{E}$ in the frame graph can be computed as follows:
\begin{equation}
    \mathbf{p}_{ij} = \Pi_c(\mathbf{G}_{ij} \circ \Pi_c^{-1}(\mathbf{p}_i, \mathbf{d}_i)), \  \mathbf{p}_{ij} \in \mathbb{R}^{H\times W \times 2}, \  \mathbf{G}_{ij} = \mathbf{G}_j \circ \mathbf{G}_i^{-1}
\end{equation}
where $\Pi_c$ is the camera model that reprojects 3D coordinate points to the image plane, while $\Pi_c^{-1}$ is the inverse function that projects the 2D coordinate grid $\mathbf{p}_i$ and the inverse depth map $\mathbf{d}$ to the 3D coordinate points. $\mathbf{G}_{ij}$ represents the relative pose of the images $I_i$ and $I_j$. $\mathbf{p}_{ij}$ is 2D coordinate grid when the coordinate of pixel $\mathbf{p}_i$ is mapped to $j$ frame with the current estimated pose and depth. The corrected correspondence represents the sum of the predicted correspondence and the optical flow residuals, i.e. $\mathbf{p}_{ij}^* = \mathbf{p}_{ij} + \mathbf{r}_{ij}$.


\noindent\textbf{DBA Layer.} We use the dense bundle adjustment layer (DBA) defined in DROID-SLAM~\cite{teed2021droid} to map stream revisions to update the current estimated pixel-wise depths and poses. The cost function can be defined as follows:
\begin{equation}
    \mathbf E(\mathbf{G}', \mathbf{d}') = \sum_{(i,j) \in \mathcal{E}} \norm{\mathbf{p}_{ij}^* - \Pi_c(\mathbf{G}'_{ij} \circ \Pi_c^{-1}(\mathbf{p}_i, \mathbf{d}'_i)) }_{\Sigma_{ij}}^2
    \label{eqn:objective}
\end{equation}
\begin{equation}
\Sigma_{ij} = \diag \mathbf{w_p}_{ij}
\end{equation}
We use the Schur complement to solve this non-linear least squares problem, Eq.~\ref{eqn:objective}. The Gauss-Newton algorithm is exploited to update the residuals of the pose ($\Delta \boldsymbol\xi$), the depth, and the mask ($\Delta \Theta$).



\begin{table*}[t]
\centering
\vspace{-1ex}
\caption{\textbf{SLAM Comparison Results on KITTI (K) \& Virtual KITTI (VK) Datasets with Metric: ATE[m].} X means system failure. }
\resizebox{\linewidth}{!}{
    \begin{tabular}{c c c c c c c c c c c c c c c c c c c c}
        \toprule
        Method & K00 & K01 & K02 & K03 & K04 & K05 & K06 & K07 & K08 & K09 & K10 & VK01 & VK02 & VK06 & VK18 & VK20 \\
        \midrule
        DynaSLAM~\cite{bescos2018dynaslam} & 8.07 & 385.33 & \underline{21.776} & \underline{0.873} & 1.402
 & \textbf{4.461} & \underline{14.364} & \textbf{2.628} & 50.369 & \underline{41.91} & \textbf{7.519} & 27.830 & X & X & X & \textbf{2.807} \\
        DROID-SLAM~\cite{teed2021droid} & \textbf{4.86} & \underline{95.45} & \textbf{18.81} & 0.893 & 0.816 & 16.03 & 42.786 & 27.402 & \underline{16.34} & 46.4 & 11.308 & \underline{1.091} & \textbf{0.025} & \underline{0.113} & \underline{1.156} & 8.285 \\
        \midrule
        Ours & \underline{5.69} & \textbf{91.19} & 23.6 & \textbf{0.855} & \textbf{0.808} & \underline{8.41} & \textbf{13.57} & \underline{8.89} & \textbf{6.67} & \textbf{14.65} & \underline{8.66} & \textbf{0.369} & \underline{0.055} & \textbf{0.113} & \textbf{0.822} & \underline{3.079} \\
        \bottomrule
    \end{tabular}
}  
    \label{tab:Kitti+vkitti}
    \vspace{-0.8em}
\end{table*}

\begin{table*}[t]
\vspace{-1ex}
	\centering
	\caption{\textbf{Absolute Trajectory Error (ATE) Comparison on TUM-RGBD Dynamic Sequences.} The best results are shown in bold. PVO achieves competitive and even best performance, outperforming DROID-SLAM in all sequences.
	}
	\label{tab:tum_ate}
	\resizebox{14cm}{!}
	{
	\begin{tabular}{cc|ccc|cc}
		\toprule
		\multicolumn{2}{c|}{\multirow{1}{*}{Sequences}}         & \multicolumn{5}{c}{ Trans. RMSE of  trajectory alignment [m]}    \\
		\multicolumn{2}{c|}{}       &  DVO SLAM~\cite{kerl2013robust} 
    &   
     ORB-SLAM2~\cite{orbslam2}  
		&  
     PointCorr~\cite{dai2020rgb} & DROID-SLAM~\cite{teed2021droid} & Ours
     \\
		\midrule
		\multirow{5}{*}{slightly dynamic}  & fr2/desk-person   &0.104 &\textbf{0.006} & \underline{0.008}  & 0.017          & 0.013  \\
		& fr3/sitting-static     &0.012      & 0.008  & 0.010  & \underline{0.007} &\textbf{0.006}   \\
		& fr3/sitting-xyz        &0.242      & \underline{0.010}  & \textbf{0.009}   &0.016  &0.014  \\
		& fr3/sitting-rpy       &0.176    & \underline{0.025}    &\textbf{0.023}   & 0.029  & 0.027   \\
		& fr3/sitting-halfsphere  &0.220   & 0.025  & \underline{0.024}  & 0.026  & \textbf{0.022}  \\
		\midrule
		\multirow{4}{*}{highly dynamic} & fr3/walking-static   &0.752 & 0.408 & \underline{0.011} & 0.016 & \textbf{0.007}  \\
		& fr3/walking-xyz      &1.383    &0.722  & 0.087  & \underline{0.019} & \textbf{0.018}  \\ 
		& fr3/walking-rpy     &1.292     &0.805 & 0.161  & \underline{0.059} &  \textbf{0.056}    \\
		& fr3/walking-halfsphere  &1.014   & 0.723   & \textbf{0.035} & 0.312  &   \underline{0.221}   \\
		\bottomrule
	\end{tabular}
	}
\vspace{-0.5em}
\end{table*}

\subsection{VO-Enhanced VPS Module}
\label{sec:vo-enhanced vps}

\begin{figure}
  \centering
  \vspace{-1em}
  \includegraphics[width=\linewidth]{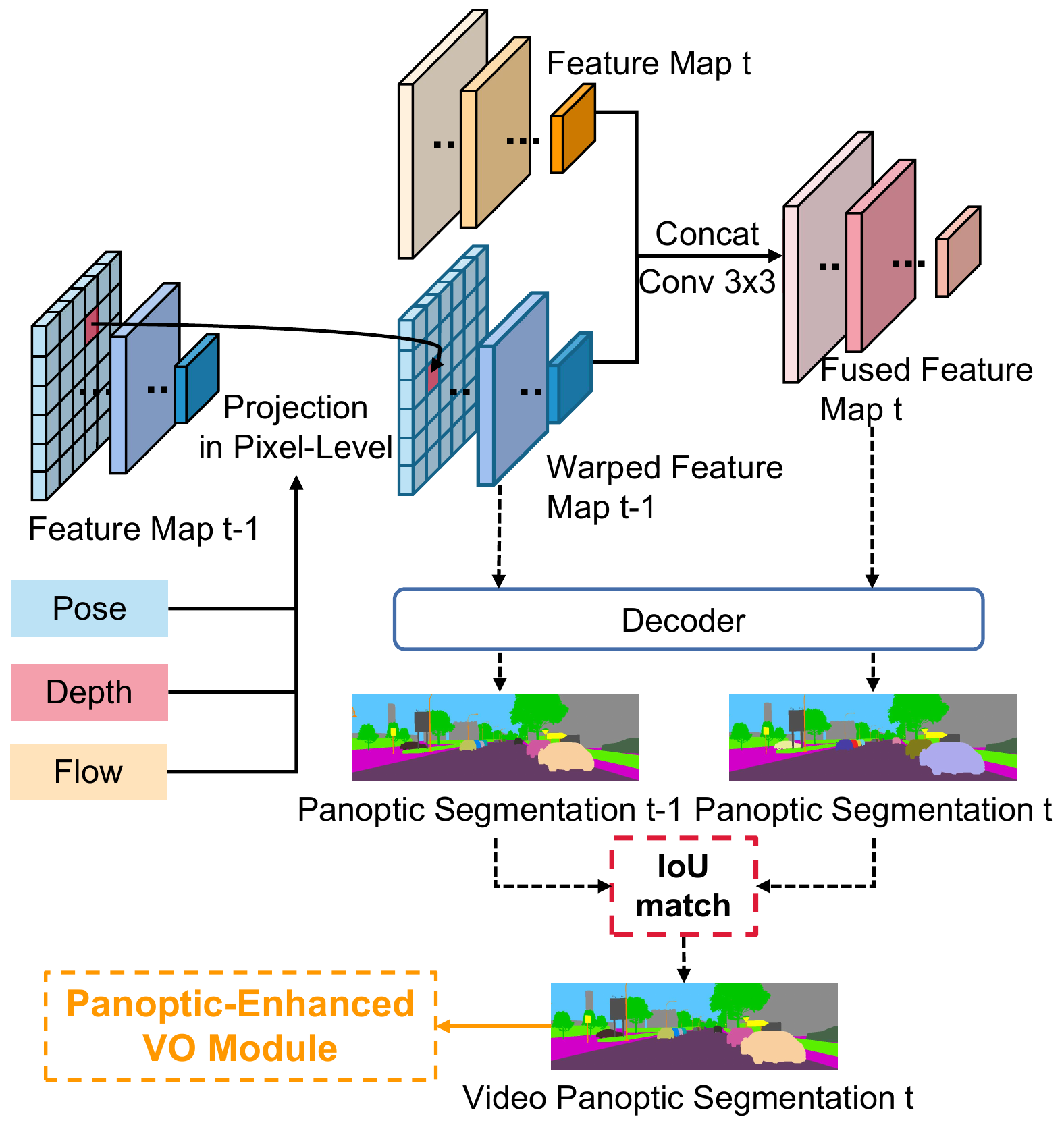}
  \caption{\textbf{VO-Enhanced VPS Module.}
    VO-Enhanced VPS Module enables feature tracking and fusion of different frames using the pose, depth, and optical flow information obtained from Visual Odometry. An online fusion module is included to better cope with occlusion challenges. The video panoptic segmentation results will be fed into the Panoptic-Enhanced VO Module.}
    \label{fig:vo-enhanced_vps_module}
\end{figure}

   
Video panoptic segmentation aims to obtain panoptic segmentation results for each frame and maintain the segmentation's consistency between frames. To improve the segmentation accuracy and tracking accuracy, some methods such as FuseTrack~\cite{kim2020video} try to use optical flow information to fuse features and track them according to the similarity of features. These methods only come from a 2D perspective that may encounter occlusion or violent motion. We live in a 3D world where additional depth information can be used to model the scene better. Our VO-Enhanced VPS Module is based on this understanding and can better solve the mentioned problems. 

Fig.~\ref{fig:vo-enhanced_vps_module} shows the VO-Enhanced VPS Module, which obtains the warped feature by warping the feature of the previous frame $t-1$ to the current frame t, using the depth, pose, and optical flow information obtained from visual odometry. An online fusion module will fuse the features of the current frame t and the warped features to obtain the fused features. 
To keep the consistency of the video segmentation, we first feed the warped features $t-1$ (containing geometric motion information) and the fused feature map $t$ into the decoder to obtain the panoptic segmentation $t-1$ and $t$, respectively. Then a simple IoU-match module is used to obtain a consistent panoptic segmentation. This result will be fed into the Panoptic-Enhanced VO Module.

\noindent \textbf{VO-Aware Online Fusion.}
\label{sec:vo-enhanced vps online_fusion}
The feature fusion network first concatenates the two features $\mathbf{z}_{t-1}$ and $\mathbf{z}_t$, and then passes through a convolutional layer with ReLU activations to obtain the fused features $\hat{\mathbf{z}}_{t}$. Inspired by NeuralBlox \cite{lionar2021neuralblox}, we propose two loss functions for supervision to ensure that online feature fusion can be effective (see Tab.~\ref{tab:vkitti_vpq}). 

\noindent \textbf{Feature Alignment Loss~\cite{lionar2021neuralblox}.} We employ a feature alignment loss to minimize the distance between $\mathbf{z_t}^{\ast}$ and $\hat{\mathbf{z_t}}$ in latent space:
\begin{equation}
  \mathcal{L}_{fea} = \big\rVert\mathbf{z_t}^{\ast} - \hat{\mathbf{z_t}} \big\rVert_1
  \label{eq:loss feature}
\end{equation}
where $\mathbf{z_t}^{\ast}$ denotes the average feature of the same pixel warped from different images to the same image.

\noindent \textbf{Segmentation Consistent Loss.} Additionally, we add a segmentation loss that minimizes the logit differences of query pixels $\mathbf{p}$ decoded using different features $\mathbf{z_t}^{\ast}$ and $\hat{\mathbf{z_t}}$:
\begin{equation}
  \mathcal{L}_{seg} =
  \sum_{\mathbf{p}\in\mathbb{P}}\big\lVert g_{\theta_d}(\mathbf{p}, \mathbf{z_t}^{\ast}) - g_{\theta_d}(\mathbf{p}, \hat{\mathbf{z_t}})\big\rVert_1
  \label{eq:loss reconstruction}
\end{equation}

\begin{table*}[]
\centering
\vspace{-1ex}
\caption{\textbf{Video Panoptic Segmentation Comparison Results on Cityscapes-VPS Validation Dataset with VO-Enhanced VPS Module Variants.} Each cell contains VPQ / VPQ\textsuperscript{Th} / VPQ\textsuperscript{St} scores. The best results are highlighted in boldface. Our method generally outperforms VPSNet-FuseTrack \cite{kim2020vps} and SiamTrack \cite{woo2021learning}.}

\resizebox{\textwidth}{!}{%
\begin{adjustbox}{max width=\textwidth}
\begin{tabular}{l|c|c|c|c| c|c}
\hline
{Methods} &\multicolumn{4}{c|}{Temporal window size} 
                & \multirow{2}{*}{VPQ} & \multirow{2}{*}{FPS} \\
\cline{2-5} {on \textbf{Cityscapes-VPS \textit{val}}} & k = 0 & k = 5 & k = 10 & k = 15 &  \\
\hline
VPSNet-Track  & 
			63.1 / 56.4 / 68.0 & 
            56.1 / 44.1 / 64.9 & 
            53.1 / 39.0 / 63.4 &
            51.3 / 35.4 / 62.9 &   
            55.9 / 43.7 / 64.8 &
            4.5\\

VPSNet-FuseTrack \quad \quad \quad   & 
            64.5 / 58.1 / 69.1 & 
            57.4 / 45.2 / 66.4 &
            54.1 / 39.5 / 64.7 &   
            52.2 / 36.0 / 64.0 & 
            57.2 / 44.7 / 66.6 &
            1.3\\

SiamTrack &
64.6 / 58.3 / 69.1 &
57.6 / 45.6 / 66.6 &
54.2 / 39.2 / 65.2 &
52.7 / 36.7 / 64.6 &
57.3 / 44.7 / 66.4 &
4.5\\
\hline
PanopticFCN~\cite{li2021panopticfcn} + Ours  & 
			\textbf{65.6} / 60.0 / 69.7 & 
            \textbf{57.8} / 45.7 / 66.6 & 
            54.3 / 39.5 / 65.1 &
            52.1 / 35.4 / 64.3 & 
            \textbf{57.5} / 45.1 / 66.4 &
            5.1 \\

VPSNet-FuseTrack + Ours \quad \quad \quad   & 
            65.0 / 59.0 / 69.4 & 
            57.6 / 45.0 / 66.7 &
            \textbf{54.4} / 39.1 / 65.6 &   
            \textbf{52.8} / 35.8 / 65.2 & 
            \textbf{57.5} / 44.7 / 66.7&
            1.1\\

\hline
\end{tabular}
\end{adjustbox}
}

\vspace{-1.1ex}
\label{tab:cityvps_vpq val and test}
\end{table*}

\begin{table*}[]
\centering
\vspace{-1.6ex}
\caption{\textbf{Video Panoptic Segmentation Comparison Results on VIPER with VO-Enhance VPS Variants.} Each cell contains VPQ / VPQ\textsuperscript{Th} / VPQ\textsuperscript{St} scores. The best results are highlighted in boldface.  Our method generally outperforms VPSNet-FuseTrack~\cite{kim2020vps} and SiamTrack~\cite{woo2021learning}.}

\resizebox{\textwidth}{!}{%

\begin{adjustbox}{max width=\textwidth}
\begin{tabular}{l|c|c|c|c| c|c}
\hline
{Methods} &\multicolumn{4}{c|}{Temporal window size} 
                & \multirow{2}{*}{VPQ} & \multirow{2}{*}{FPS} \\
 \cline{2-5} {on \textbf{VIPER}} & k = 0 & k = 5 & k = 10 & k = 15 &  \\
\hline
VPSNet-Track  & 
		    48.1 / 38.0 / 57.1 & 
            49.3 / 45.6 / 53.7 & 
            45.9 / 37.9 / 52.7 &
            43.2 / 33.6 / 51.6 &  
            46.6 / 38.8 / 53.8 &
            5.1 \\
            
VPSNet-FuseTrack      & 
			49.8 / 40.3 / 57.7 & 
            51.6 / 49.0 / 53.8 & 
            47.2 / 40.4 / 52.8 &
            45.1 / 36.5 / 52.3 &   
            48.4 / 41.6 / 53.2 &
            1.6 \\
            SiamTrack &
            51.1 / 42.3 / 58.5 &
            \textbf{53.4} / 51.9 / 54.6 &
            49.2 / 44.1 / 53.5 &
            47.2 / 40.3 / 52.9 &
            50.2 / 44.7 / 55.0 &
            5.1 \\
\hline

 PanopticFCN + Ours  & 
			\textbf{54.6} / 50.3 / 57.9 & 
            51.7 / 44.5 / 57.3 & 
            \textbf{50.5} / 41.8 / 57.2 &
            \textbf{49.1} / 38.9 / 56.9 &     
            \textbf{51.5} / 43.9 / 57.3 &
            3.6 \\

\hline
\end{tabular}
\end{adjustbox}
}
\vspace{-1ex}
\label{tab:viper_vpq}
\end{table*}

\subsection{Recurrent Iterative Optimization}
We can optimize the proposed Panoptic-Enhanced VO Module and VO-Enhanced VPS Module in a recurrent iterative manner until convergence, which is inspired by the EM algorithm. Experimentally, it generally takes only two iterations for the loop to converge. Tab.~\ref{tab:vkitti_vpq} and Tab.~\ref{table:vkitti Panoptic-VO Module} demonstrate that recurrent iterative optimization can boost the performance of both the VPS and VO Modules.

\subsection{Implementation Details}
Implemented by PyTorch, PVO consists of three main modules: image panoptic segmentation, Panoptic-Enhanced VO Module, and VO-Enhanced VPS Module. We use three stages to train our network. Image panoptic segmentation is trained on Virtual KITTI~\cite{cabon2020virtual} dataset as initialization. Following PanopticFCN, we adopt a multi-scale scaling policy during training. We optimize the network with an initial rate of 1e-4 on two GeForce RTX 3090 GPUs, where each mini-batch has eight images. The SGD optimizer is used with a weight decay of 1e-4 and momentum of 0.9. The training of the Panoptic-Enhanced VO Module follows DROID-SLAM~\cite{teed2021droid}, except that it additionally feeds the ground-truth panoptic segmentation results. Specifically, we trained this module on the Virtual KITTI dataset with two GeForce RTX-3090 GPUs for 80,000 steps, which took about two days. When training the VO-enhanced video panoptic segmentation module, we use the ground-truth depth, optical flow, and pose information as geometric priors to align the features, and fix the backbone of the trained single-image panoptic segmentation, and then train the fusion module only. The network is optimized with an initial learning rate of 1e-5 on one GeForce RTX 3090 GPU, where each batch has eight images. When the fusion network has largely converged, we add a segmentation consistency loss function to refine our VPS Module further.

    

\begin{table*}[]
\centering
\vspace{-1ex}
\caption{\textbf{Ablation Study of VO-Enhanced VPS Module Variants on VKITTI2 Dataset.} Each cell contains VPQ / VPQ\textsuperscript{Th} / VPQ\textsuperscript{St} scores. The best results are highlighted in boldface. Our method performs better than existing video panoptic segmentation methods.}

\resizebox{\textwidth}{!}{%

\begin{adjustbox}{max width=\textwidth}
\begin{tabular}{l|c|c|c|c| c}
\hline
{Methods} &\multicolumn{4}{c|}{Temporal window size} 
                & \multirow{2}{*}{VPQ} \\
 \cline{2-5} {on \textbf{VKITTI2}} & k = 0 & k = 5 & k = 10 & k = 15 &  \\
\hline
VPS baseline  & 
		    58.24 / 60.11 / 57.93 & 
            55.50 / 53.78 / 56.28 & 
            54.13 / 50.29 / 55.53 &
            53.65 / 48.53 / 55.46 &  
            54.90 / 51.95 / 56.05 \\
VPS baseline + w/fusion  & 
		    59.16  / 67.00 / 56.91 & 
            56.27 / 60.98 / 54.96 & 
            54.96 / 57.74 / 54.18 &
            54.58 / 55.97 / 54.19 &  
            55.81 / 59.23 / 54.85  \\
            \hline
Ours (VO-$>$VPS + w/o fusion)  & 
		    58.24  / 60.11 / 57.93 & 
            55.67 / 54.44 / 56.28 & 
            54.29 / 50.91 / 55.53 &
            53.83 / 49.22 / 55.46 &  
            55.04 / 52.48 / 56.05  \\
Ours (VO-$>$VPS + w/fusion + w/o fea loss)  & 
		    58.51 / 64.07 / 56.97 & 
            55.62 / 58.53 / 54.86 & 
            54.29 / 55.15 / 54.13 &
            53.94 / 53.40 / 54.19 &  
            55.14  / 56.62 / 54.81   \\
Ours (VO-$>$VPS + w/fusion + w/o seg loss)  & 
		    58.73 / 65.05 / 56.95 & 
            55.83 / 59.34 / 54.89 & 
            54.51 / 56.01 / 54.15 &
            54.15 / 54.26 / 54.19 &  
            55.37 / 57.49 / 54.82   \\
\hline
Ours (VO-$>$VPS)  & 
		    59.18  / 67.00 / 56.94 & 
            56.25 / 61.00 / 54.93 & 
            54.94 / 57.77 / 54.15 &
            54.57 / 56.01 / 54.17 &  
            55.80 / 59.25 / 54.83  \\
Ours (VO-$>$VPS + w/o depth ) x2  & 
		    59.17 / 66.87 / 56.95 & 
            56.39 / 61.45 / 56.25 & 
            55.04 / 58.15 / 54.15 &
            54.72 / 56.46 / 54.22 &  
            55.89 / 59.57 / 54.83  \\
            
Ours (VO-$>$VPS) x2  & 
		    \textbf{59.18}  / 67.00 / 56.94 & 
            \textbf{56.42} / 61.67 / 54.93 & 
            \textbf{55.10} / 58.40 / 54.15 &
            \textbf{54.84} / 56.67 / 54.17 &  
            \textbf{55.94} / 59.77 / 54.83  \\

\hline
\end{tabular}
\end{adjustbox}
}
\vspace{-1ex}
\label{tab:vkitti_vpq}
\end{table*}

\begin{table}[h]
\vspace{-2ex}
\caption{\textbf{Ablation Study of Panoptic-Enhanced VO Module Results on VKITTI2 Dataset.} Our method outperforms DROID-SLAM on most of the highly dynamic VKITTI2 datasets, and the accuracy of the pose estimation is significantly improved and slightly slowed down after recurrent iterative optimization.}
\resizebox{1.0\linewidth}{!}{%
\begin{tabular}{l|cccccccc | c}
\toprule
Monocular & 01 & 02 & 06 & 18 & 20 & Avg \\
\midrule
DROID-SLAM~\cite{teed2021droid} & 1.091 & \textbf{0.025} & 0.113 & 1.156 & 8.285 & 2.134 \\
Ours (VPS-$>$VO w/o filter) & 0.384 & 0.061 & 0.116 & 0.936 & 5.375 & 1.374 \\
Ours (VPS-$>$VO) & 0.374 & 0.057 & 0.113 & 0.960 & 3.487 & 0.998 \\
Ours (VPS-$>$VO x2) & 0.371 & 0.057 & 0.113 & 0.954 & 3.135 & 0.926\\
Ours (VPS-$>$VO x3)  & \textbf{0.369} & 0.055 & \textbf{0.113} & \textbf{0.822} & \textbf{3.079} & \textbf{0.888}\\
\midrule
DROID-SLAM's runtime (FPS)  & 5.73 & 12.67 & 19.96 & 7.08 & 10.20 & 11.13 \\
Ours' runtime (FPS)  & 4.45 & 9.69 & 14.52 & 6.22 & 8.10 & 8.60 \\
\midrule
\end{tabular}
\label{table:vkitti Panoptic-VO Module}
}
\vspace{-1ex}
\end{table}

\vspace{-1em}
\section{Experiments}
For visual odometry, we conduct experiments on three datasets with dynamic scenes: Virtual KITTI, KITTI, and TUM RGBD dynamic sequences. Absolute Trajectory Error (ATE) is used for evaluation. For video panoptic segmentation, we use Video Panoptic Quality (VPQ) metric~\cite{kim2020video} on Cityscapes and VIPER datasets. We further perform ablation studies on Virtual KITTI to analyze the design of our framework. Finally, we demonstrate the applicability of our PVO on video editing, as shown in Sec.~\ref{sec: video editing} in the supplementary materials.


\subsection{Visual Odometry}

\begin{figure}[h]
  \centering
  \includegraphics[width=\linewidth]{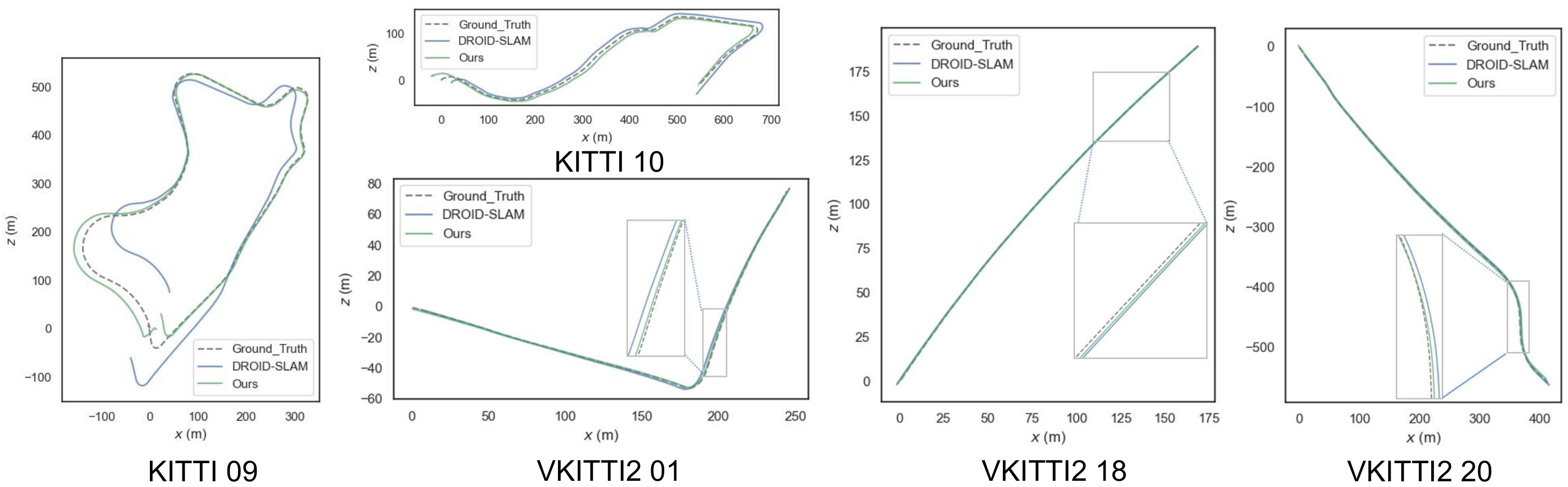}
  \caption{\textbf{Trajectory Comparison on KITTI and VKITTI2.} Our method performs better than DROID-SLAM, having better trajectory estimation results.}
\label{fig: trajetory of vkitti2}
\vspace{-2ex}
\end{figure}

\noindent \textbf{VKITTI2.} Virtual KITTI dataset~\cite{cabon2020virtual} consists of 5 sequences cloned from the KITTI tracking benchmark, which provides 
RGB, depth, class segmentation, instance segmentation, camera pose, flow, and scene flow data 
for each sequence. 
As shown in Tab.~\ref{table:vkitti Panoptic-VO Module} and Fig.~\ref{fig: trajetory of vkitti2}, our PVO outperforms DROID-SLAM by a large margin for most sequences and achieves competitive performance in sequence 02. 

\noindent \textbf{KITTI.} KITTI~\cite{geiger2013vision} is a dataset capturing real-world traffic scenarios, ranging from freeways over rural areas to urban streets with plenty of static and dynamic objects. 
We applied the PVO model trained on the VKITTI2 ~\cite{cabon2020virtual} dataset to the KITTI~\cite{geiger2013vision} sequences. As shown in Fig.~\ref{fig: trajetory of vkitti2} (KITTI 09 and 10 sequences), the pose estimation error of PVO is only half that of DROID-SLAM, which proves the good generalization ability of PVO. 
Tab.~\ref{tab:Kitti+vkitti} shows the complete SLAM comparison results on KITTI and VKITTI datasets, where PVO outperforms DROID-SLAM and DynaSLAM by a large margin in most scenarios. Note that we use the code of DynaSLAM, which is a classic SLAM system with instance segmentation. DynaSLAM falls into the catastrophic system failure in the VKITTI2 02, 06, and 18 sequences.



\noindent \textbf{TUM-RGBD.} TUM RGBD is a dataset capturing indoor scenes with a handheld camera. We choose the dynamic sequences of the TUM RGBD dataset to show the effectiveness of our method. We compare PVO with DROID-SLAM and three state-of-the-art dynamic RGB-D SLAM systems, namely DVO SLAM~\cite{kerl2013robust}, ORB-SLAM2~\cite{orbslam2} and PointCorr~\cite{dai2020rgb}. Note that PVO and DROID-SLAM only use monocular RGB videos. Tab.~\ref{tab:tum_ate} demonstrates that PVO outperforms DROID-SLAM in all scenes. Compared to the conventional RGB-D SLAM systems, our method also performs better in most of the scenes.



\subsection{Video Panoptic Segmentation}
We compare PVO with three instance-based video panoptic segmentation methods, namely VPSNet-Track, VPSNet-FuseTrack~\cite{kim2020vps}, and SiamTrack~\cite{woo2021learning}. Built on the image panoptic segmentation model UPSNet~\cite{xiong2019upsnet}, VPSNet-Track additionally adds MaskTrack head~\cite{yang2019video} to form the video panoptic segmentation model. 
VPSNet-FuseTrack based on VPSNet-Track additionally injects temporal feature aggregation and fusion. While SiamTrack finetunes VPSNet-Track with the pixel-tube matching loss~\cite{woo2021learning} and the contrast loss and has slight performance improvement. VPSNet-FuseTrack is mainly compared because the code of SiamTrack is not available.

\noindent \textbf{Cityscapes.} We adopt the public train/val/test split of Cityscapes in VPS~\cite{kim2020video},  where each video contains 30 consecutive frames, with the corresponding ground truth annotations for every five frames.
Tab.~\ref{tab:cityvps_vpq val and test} demonstrates that our method with PanopticFCN~\cite{li2021panopticfcn} outperforms the state-of-the-art method on the val dataset, achieving \textbf{+1.6\% VPQ} higher than the VPSNet-Track. Compared with VPSNet-FuseTrack~\cite{kim2020video}, our method has slight improvement and can keep consistent video segmentation, shown in Fig.~\ref{fig:qualtitative result compared with vpsnet} in the supplementary materials. The reason is that our VO Module only obtains 1/8 resolution optical flow and depth due to the limited memory.




\noindent \textbf{VIPER.} VIPER maintains plenty of high-quality panoptic video annotations, which is another video panoptic segmentation benchmark. We follow VPS~\cite{kim2020vps} and adopt its public train/val split. We use 10 selected videos from day scenarios and the first 60 frames of each video are used for evaluation. Tab.~\ref{tab:viper_vpq} demonstrates that compared with VPSNet-FuseTrack, our method with PanopticFCN achieves much higher scores (\textbf{+3.1 VPQ}) on the VIPER dataset.


\subsection{Ablation Study}
\begin{figure}
  \centering
  \vspace{-1ex}
  \includegraphics[width=0.9\linewidth]{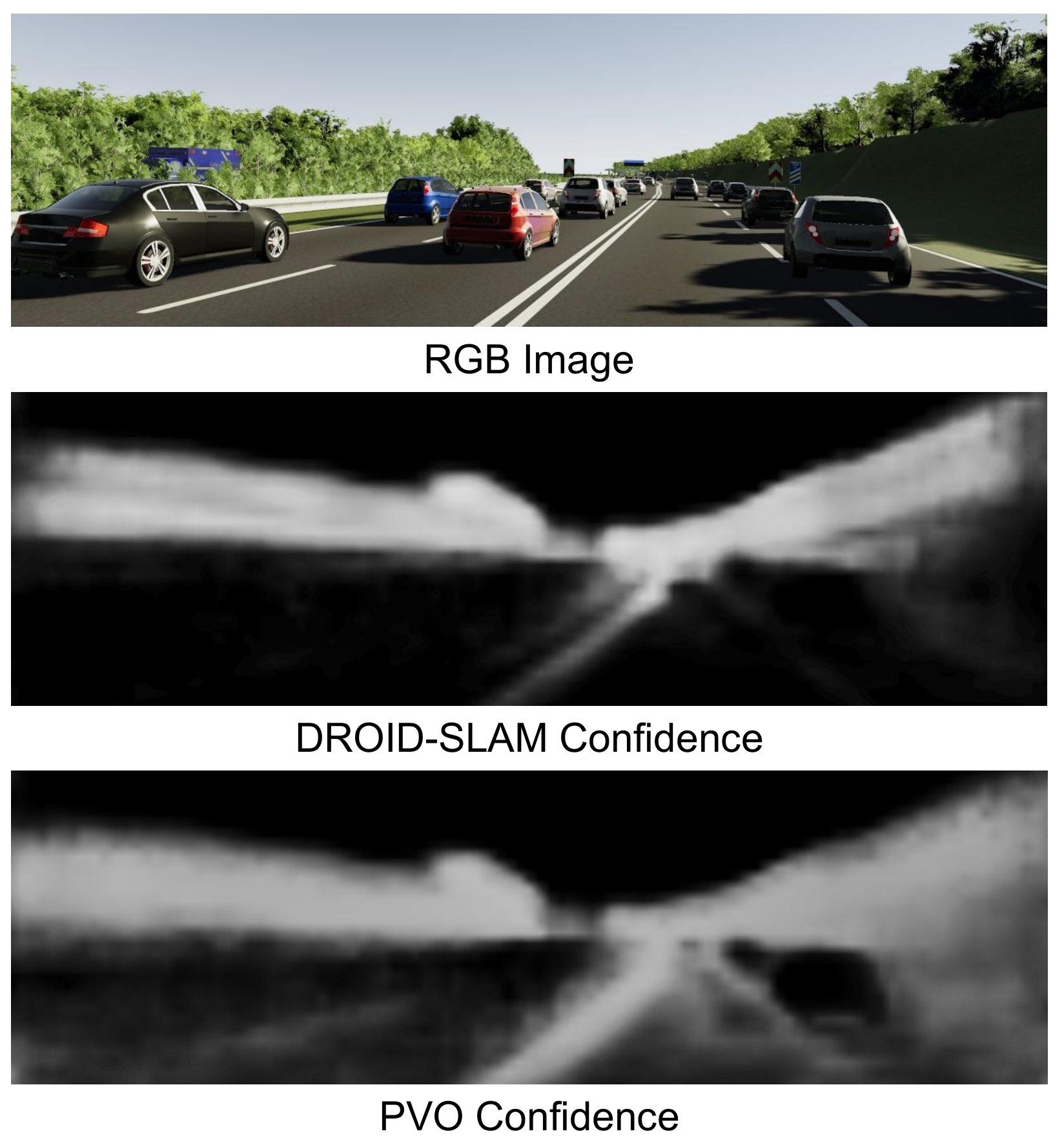}
  \caption{\textbf{Panoptic-Aware Confidence.} We visualize the confidence of the PVO model vs. DROID-SLAM. We can see that with panoptic information, the panoptic weights can better remove the dynamic interference and keep the static features for solving the camera pose. The black color indicates that the confidence tends to be close to 0. }
    \label{fig:Panoptic-Aware Weight}
  \vspace{-0.8em}
\end{figure}

\noindent \textbf{VPS-Enhanced VO Module.} In the Panoptic-Enhanced VO Module, we use DROID-SLAM~\cite{teed2021droid} as our baseline. (VPS-$>$VO) means the panoptic information prior was added to enhance the VO baseline. (VPS-$>$VO x2) means that we can iteratively optimize the VO Module twice. (VPS-$>$VO x3) means recurrent iterative optimization on the VO Module 3 times. Tab.~\ref{table:vkitti Panoptic-VO Module} and Fig.~\ref{fig:Panoptic-Aware Weight} show the panoptic information can help improve the accuracy of DROID-SLAM on most of the highly dynamic VKITTI2 datasets. The recurrent iterative optimization can further improve the results.

\noindent \textbf{VO-Enhanced VPS Module.} 
To evaluate whether VO helps VPS, we first use PanopticFPN~\cite{kirillov2019panoptic} to get the panoptic segmentation results for each frame, and then use the optical flow information from RAFT~\cite{raft} for inter-frame tracking. This is set as VPS baseline. (VPS baseline + w/fusion) means we additionally fuse the feature with the flow estimation. (VO-$>$VPS + w/o fusion) means that we use additional depth, pose, and other information on top of the baseline. (VO-$>$VPS) means we additionally fuse the feature. (VO-$>$VPS x2) means that we use the recurrent iterative optimization module to enhance the VPS results further.
As shown in Tab.~\ref{tab:vkitti_vpq} 
and Fig.~\ref{fig:qualtitative result compared with baseline on kitti2} in the supp. materials, the VO-Enhanced VPS Module is effective in improving segmentation accuracy and tracking consistency. 

\noindent \textbf{Online Fusion in VO-Enhanced VPS Module.}
To validate the effectiveness of the proposed Feature Alignment Loss (fea loss) and Segmentation Consistent Loss (seg loss), the methods are followed:
(VO-$>$VPS + w/fusion + w/o fea loss) means that we train the online fusion module without Feature Alignment Loss.
(VO-$>$VPS + w/fusion + w/o seg loss) means that we train the online fusion module without Segmentation Consistent Loss.
Tab.~\ref{tab:vkitti_vpq} demonstrates the effectiveness of these two loss function.

\vspace{-0.7em}
\section{Conclusion}
We have presented a novel panoptic visual odometry method, which models the VO and the VPS in a unified view, enabling the two tasks to facilitate each other. The panoptic update module can help improve the pose estimation, while the online fusion module helps improve the panoptic segmentation. Extensive experiments demonstrate that our PVO outperforms state-of-the-art methods in both tasks. 

\noindent \textbf{Limitations.} The main limitation is that PVO is built on DROID-SLAM and panoptic segmentation, which makes the network heavy and requires much memory. 
Although PVO can perform robustly in dynamic scenes, it ignores the problem of loop closure when the camera returns to the previous position. 
Exploring a low-cost and efficient SLAM system with loop closure is our future work.

\section{Acknowledgements} 
This work was partially supported by NSF of China (No. 61932003) and ZJU-SenseTime Joint Lab of 3D Vision. Weicai Ye was partially supported by China Scholarship Council (No. 202206320316).

{\small
\bibliographystyle{ieee_fullname}
\bibliography{PVO}
}

\clearpage
\appendix


\twocolumn[
    \centering
    \Large
    \textbf{PVO: Panoptic Visual Odometry} \\
    \vspace{0.5em}Supplementary Material \\
    \vspace{1.0em}
] 


\setcounter{table}{0}
\setcounter{figure}{0}
\setcounter{equation}{0}
\renewcommand{\thetable}{\thesection\arabic{table}}
\renewcommand{\thefigure}{\thesection\arabic{figure}}
\renewcommand{\theequation}{\thesection\arabic{equation}}

In this supplementary document, we provide more experiment results (Sec.~\ref{sec:Experiments Results}), such as the ablation study of our method. We further demonstrate the applicability of our method in video editing (Sec.~\ref{sec: video editing}) and discuss the limitation (Sec.~\ref{sec:discussion}) of PVO in video editing. We also provide the supplementary video which demonstrates the qualitative results of our method and the video editing effects.
 
\section{Experiments Results}
\label{sec:Experiments Results}

\begin{table*}[t]
\caption{\textbf{Ablation Study of Panoptic-Enhanced VO Module on Virtual KITTI2 Dataset.} Panoptic-Enhanced VO Module outperforms DROID-SLAM on most of the highly dynamic VKITTI2 datasets, and the accuracy of the pose estimation is significantly improved after recurrent iterative optimization. The dynamic threshold set as 0.5 can achieve the best performance.}
\resizebox{0.8\linewidth}{!}{%
\begin{tabular}{l|cccccccc | c}
\toprule
Monocular & vkitti01 & vkitti02 & vkitti06 & vkitti18 & vkitti20 & Avg \\
\midrule
DROID-SLAM & 1.091 & \textbf{0.025} & 0.113 & 1.156 & 8.285 & 2.134 \\
Ours (VPS-$>$VO w/o filter) & 0.384 & 0.061 & 0.116 & 0.936 & 5.375 & 1.374 \\
Ours (VPS-$>$VO) & 0.374 & 0.057 & 0.113 & 0.960 & 3.487 & 0.998 \\
Ours (VPS-$>$VO x2) & 0.371 & 0.057 & 0.113 & 0.954 & 3.135 & 0.926\\
Ours (VPS-$>$VO x3)  & \textbf{0.369} & 0.055 & 0.113 & \textbf{0.822} & \textbf{3.079} & \textbf{0.888}\\
Ours (VPS-$>$VO x3) threshold=0.1  & 0.377 & 0.052 & \textbf{0.112} & 0.950 & 3.240 & 0.946\\
Ours (VPS-$>$VO x3) threshold=0.3  & 0.374 & 0.054 & 0.113 & 0.946 & 3.107 & 0.919\\
Ours (VPS-$>$VO x3) threshold=0.5 & 0.369 & 0.055 & 0.113 & 0.822 & 3.079 & 0.888\\
Ours (VPS-$>$VO x3) threshold=0.7  & 0.384 & 0.059 & 0.114 & 0.863 & 22.993 & 4.883\\
Ours (VPS-$>$VO x3) threshold=0.9  & 1.348 & 0.065 & 0.119 & 0.885 & 17.337 & 3.951 \\

\midrule
\end{tabular}
\label{table:vkitti ablation study Panoptic-VO Module supp}
}
\end{table*}

\subsection{Ablation Study of Panoptic-Enhanced VO Module}
\begin{figure}
  \centering
  \includegraphics[width=\linewidth]{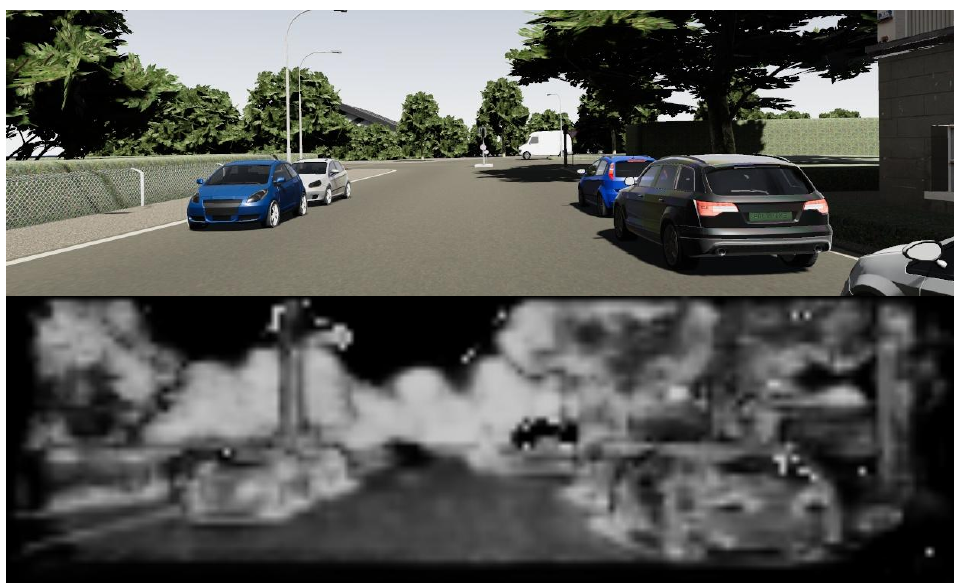}
   \caption{\textbf{Dynamic Probability of Parked Cars.} 
   The black color indicates that the confidence tends to be close to 0.}
   \label{fig: parked}
\end{figure}

In our Panoptic-Enhanced VO Module, unlike DROID-SLAM~\cite{teed2021droid}, we adjust the confidence by incorporating information from the panoptic segmentation. The dynamic mask is adjusted to the panoptic-aware dynamic mask given the initialized panoptic segmentation. Panoptic segmentation treats trees and buildings as stuff (i.e., the background is static), people and cars, etc. as things (i.e., the foreground). So the foreground objects with a high probability of motion are set to dynamic. We show an example of waiting for a traffic light in Fig.~\ref{fig: parked}, where the white color indicates that the parked cars are static.
We find the dynamic threshold set as 0.5 may achieve the best results, shown in Tab.~\ref{table:vkitti ablation study Panoptic-VO Module supp}. The reason is that when the dynamic threshold is small, too many static pixel points may be removed, while the dynamic threshold is too large, and small movements may be ignored. 
The confidence and panoptic-aware dynamic mask are passed through a panoptic-aware filter module to obtain the panoptic-aware confidence. As shown in Tab.~\ref{table:vkitti ablation study Panoptic-VO Module supp}, the panoptic-aware filter module can help improve the estimation of camera pose. 

\begin{figure}
  \centering
  \includegraphics[width=\linewidth]{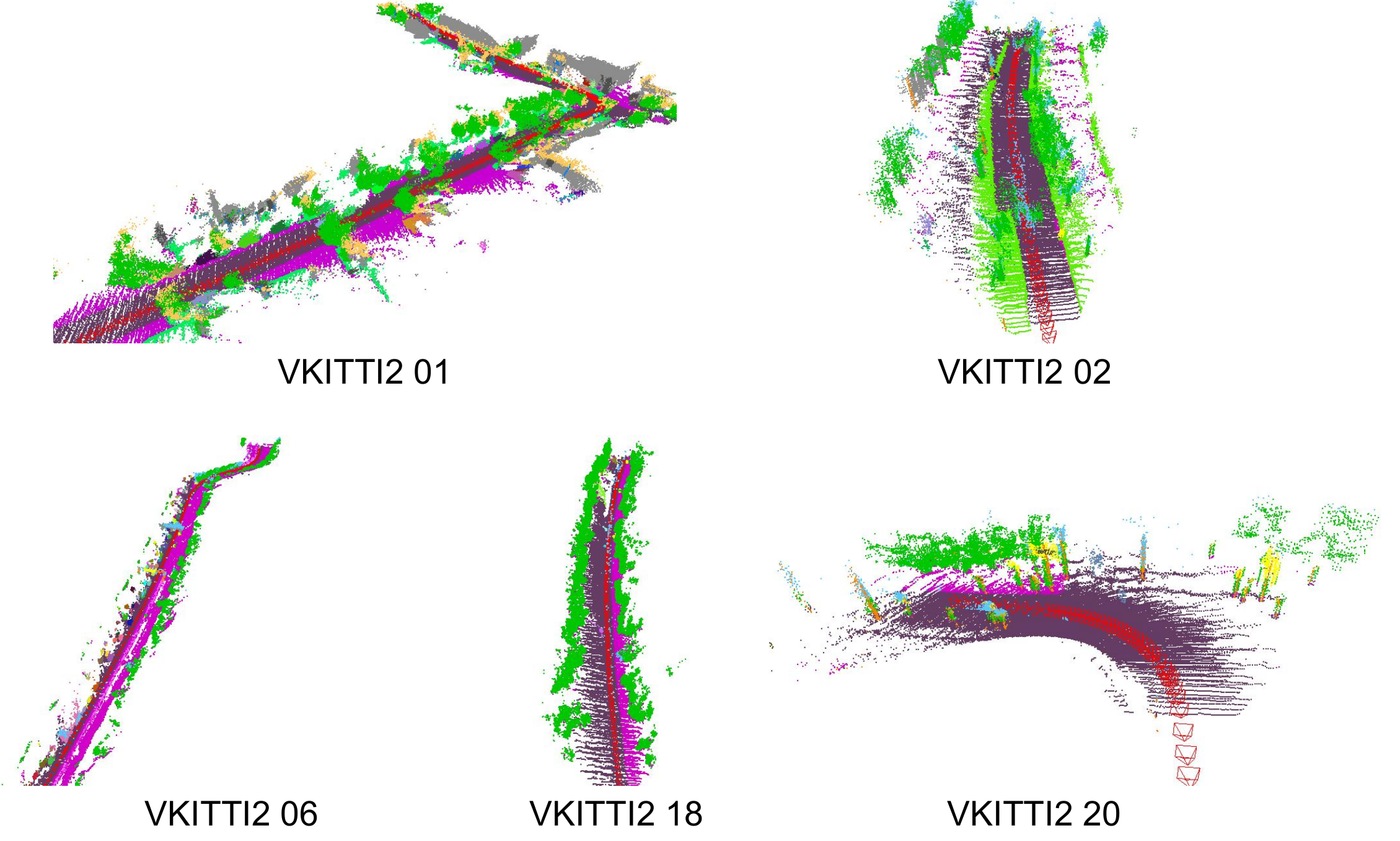}
  \caption{\textbf{Qualitative Results of Panoptic 3D Map Produced by PVO on Virtual KITTI Dataset.} We show the panoptic 3D map produced by our method. The red triangles indicate the camera pose, and different colors indicate different instances.}
\label{fig:paoptic3d map}
\end{figure}

We show the qualitative results of the panoptic 3D maps produced by our method, shown in Fig.~\ref{fig:paoptic3d map}. The supplementary video also shows how our method works.


\begin{figure}
  \centering
  \includegraphics[width=\linewidth]{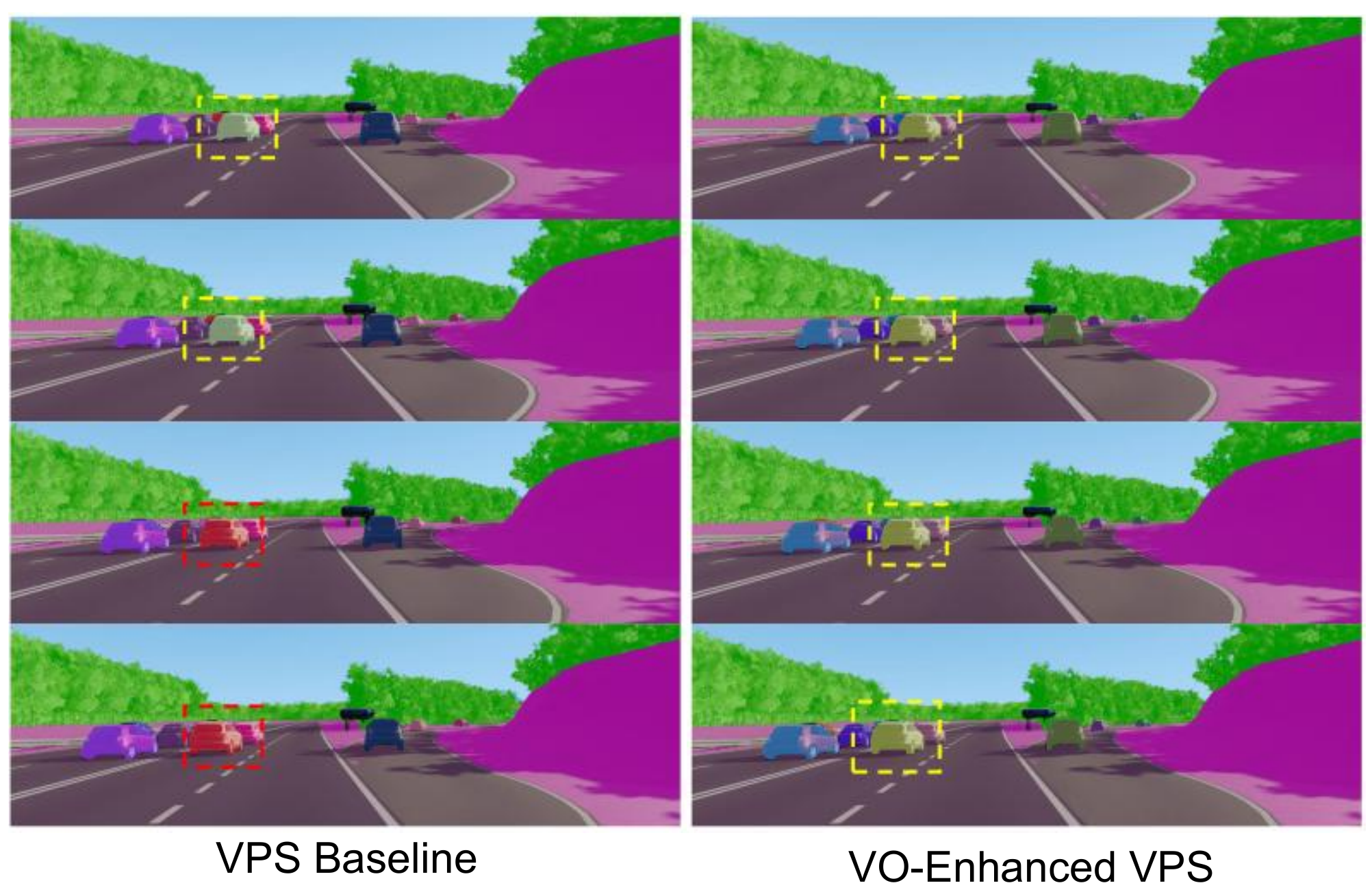}
  \vspace{-1em}
  \caption{\textbf{Comparison Results of Our VO-Enhanced VPS Module with VPS Baseline on VKITTI2 Dataset}. Our method keeps the consistent video segmentation for it is better to cope with occlusion. Different colors indicate tracking failure.}
    \label{fig:qualtitative result compared with baseline on kitti2}
\end{figure}

\begin{figure}
\centering
    \includegraphics[width=\linewidth]{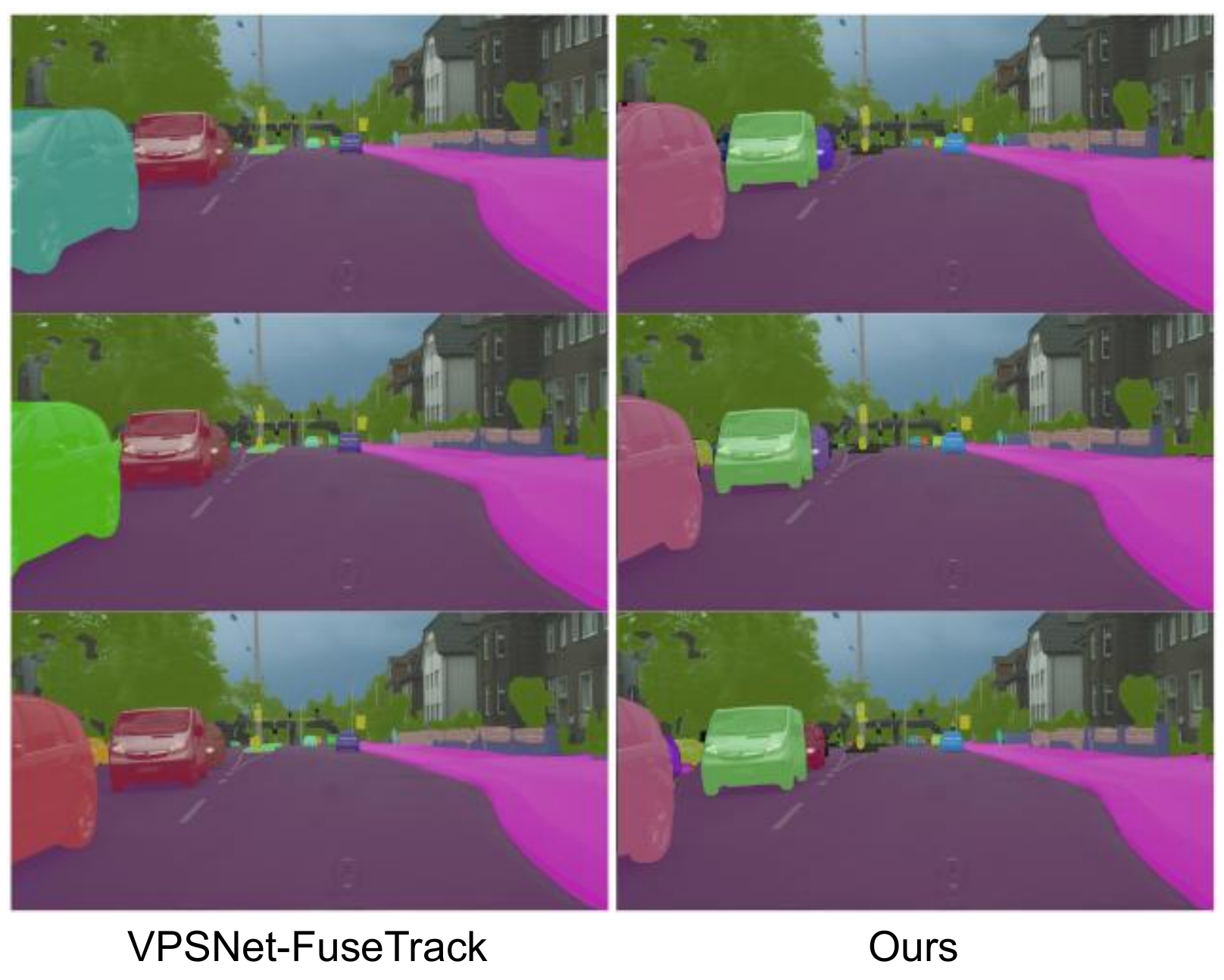}
    \vspace{-2em}
    \caption{\textbf{Comparison Results of Our method with VPSNet-FuseTrack \cite{kim2020video} on Cityscape-VPS Val Dataset.} Compared with VPSNet-FuseTrack, our method can keep consistent video segmentation. Different colors indicate tracking failure.}
  \label{fig:qualtitative result compared with vpsnet}
\end{figure}


\subsection{Ablation Study of VO-Enhanced VPS Module}
Qualitative results in Fig.~\ref{fig:qualtitative result compared with baseline on kitti2} shows our method can cope with occlusion better on VKITTI2 dataset. Fig.~\ref{fig:qualtitative result compared with vpsnet} demonstrates our method keeps consistent video panoptic segmentation on Cityscape-VPS dataset, compared with VPSNet-FuseTrack.

\section{Video Editing Applications with PVO}
\label{sec: video editing} 

\begin{figure*}[t]
  \centerline{
\begin{tabular}{ccc}
\includegraphics[width=0.333\textwidth]{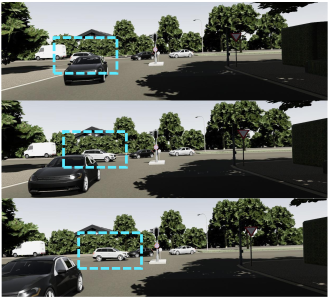} &
\includegraphics[width=0.333\textwidth]{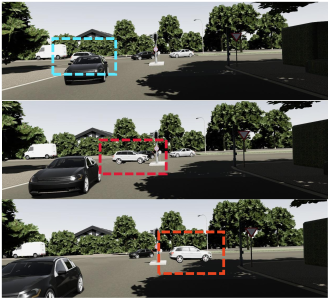} &
\includegraphics[width=0.333\textwidth]{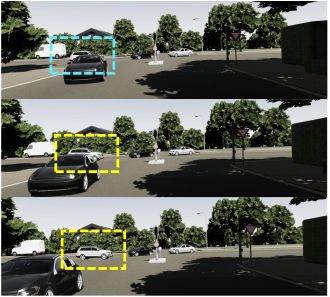} \\
(a) Original video & (b) Speed up & (c) Slow down \\
\includegraphics[width=0.333\textwidth]{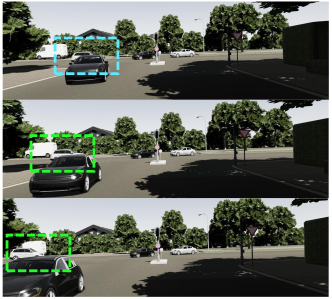} &
\includegraphics[width=0.333\textwidth]{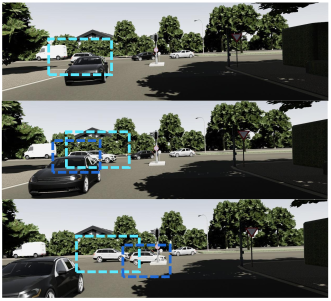} &
\includegraphics[width=0.333\textwidth]{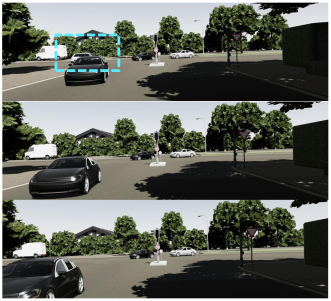} \\
(d) Reverse & (e) Copy \& paste & (f) Delete \\
\end{tabular}
}
  \caption{\textbf{Panoptic Visual Odometry (PVO) can Support Many Video Editing Effects of Motion Control.} With PVO, we can manipulate the white car in the original video with different motions and keep the overall consistency of the video. (a) The car in the original video; (b) Speed the car up; (c) Slow the car down, (d) Put the car in reverse, (e) Copy the new similar car keeping the similar motion, (f) Delete the car. The cyan box indicates the original motion pattern, and the other colors indicate our motion manipulation effect.}
  \label{fig:teaser_edit}
\end{figure*}

\begin{figure}[t]
  \centering
  \includegraphics[width=1.0\linewidth]{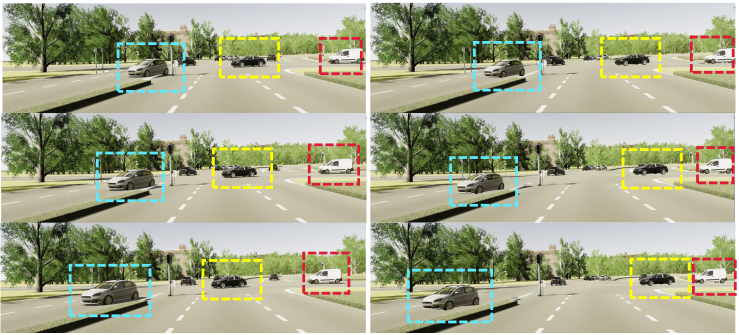}
  \caption{\textbf{Multi-Instance Motion Control.} PVO allows a more comprehensive modeling of the scene's motion, panoptic segmentation, and geometric information. PVO can support different motion manipulation of multiple moving objects, even if the camera is also moving. The blue box indicates accelerating the car, the yellow box indicates reversing, and the red box indicates decelerating the car.}
  \label{fig:motion_controler}
\end{figure}

\begin{figure}[t]
  \centering
  \vspace{-1em}
  \includegraphics[width=\linewidth]
  {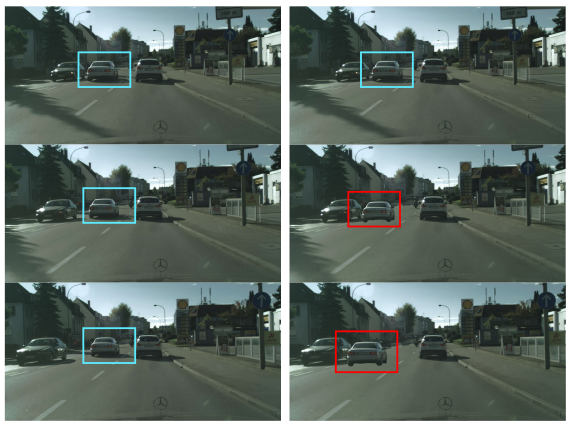}
  \caption{\textbf{Generalization Results of Motion Control on Cityscapes Dataset.} PVO demonstrates generalizability in natural scenes.}
  \label{fig:general motion}
\end{figure}

\begin{figure*}[t]
  \centering
  \includegraphics[width=\linewidth]{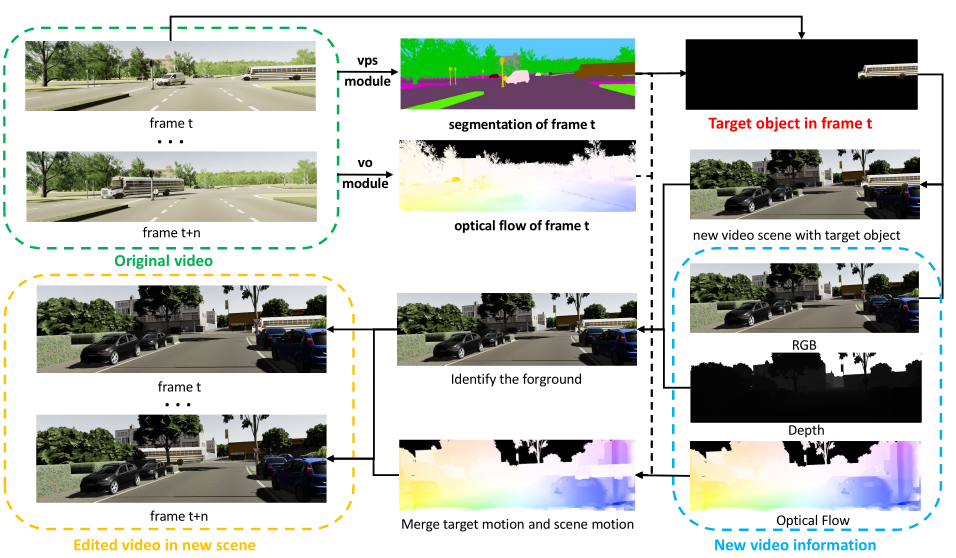}
  \vspace{-1em}
  \caption{\textbf{Video Editing Pipeline with Panoptic Visual Odometry.} PVO comprehensively models the panoptic segmentation and motion information of the entire scene. The motion of dynamic objects can be decomposed into static fields and dynamic fields. We can select an instance and directly manipulate its static fields and dynamic fields to generate a new video. Moreover, the original moving objects can be inserted into the new scene. Some of the occlusion completion and depth checks are taken into account to create a more realistic editing effect.}
\label{fig:video_editing_illustration}
\end{figure*}

\begin{figure*}[t]
\centerline{
\begin{tabular}{ccc}
\includegraphics[width=0.30\textwidth]{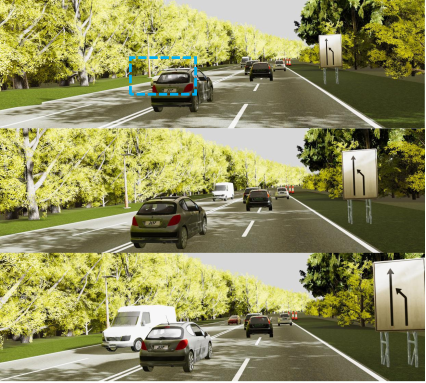} &
\includegraphics[width=0.30\textwidth]{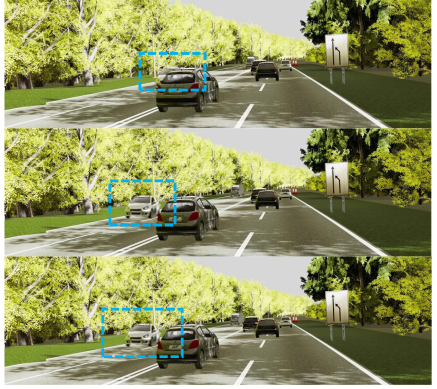} &
\includegraphics[width=0.30\textwidth]{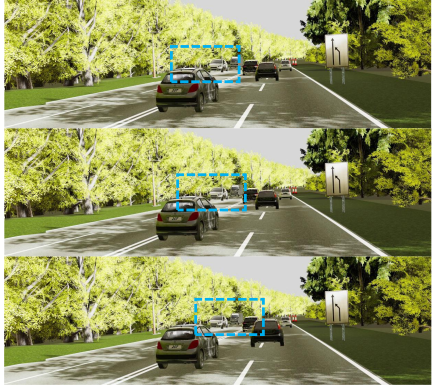} \\
(a) Original video   & (b) Pulling cars left and right   & (c) Pulling cars back and forth \\
\end{tabular}
}
  \caption{\textbf{Multi-Object Occlusion Interaction.} PVO can model the motion, panoptic segmentation, and geometric information of the scene more comprehensively. PVO can support completing multiple mutually occluding moving objects. From left to right, they are a: the original video, b: the occluded part can be restored by pulling the car left and right. c: the occluded part can be restored by pulling the car back and forth.}
  \label{fig:multi-object_interation}
\end{figure*}




In this section, we show the applicability of video editing with PVO, as shown in Fig.~\ref{fig:teaser_edit}. We can obtain rich 2D and 3D information from panoptic visual odometry, which can be utilized in video editing.

Fig.~\ref{fig:video_editing_illustration} illustrates how we can perform consistent video editing using panoptic visual odometry. Firstly, we feed the original video frames from t to t+n into the PVO network. The VO-Enhanced VPS Module and VPS-Enhanced VO Module will get the panoptic segmentation result and optical flow estimation, depth, and pose information for each frame. In addition, the motion of dynamic objects can be decomposed into the dynamic field and static field of the camera. Similarly, the above operation in the new scene can get the whole scene modeling information. We can first select one instance of the original video, then obtain the motion of the target in the new scene by merging the static field of the new scene and the dynamic field of the selected object of the original video, together with additional information such as depth checks and occlusion completion, to complete the video effect of inserting the object into the new scene. In this way, we can perform several vivid video effects, including motion control, replication, deletion, and instance interaction. Note that when the initial segmentation is incomplete, with PVO, we can first fill in the occluded parts from multiple views, thus ensuring the integrity of the object, as shown in Fig.~\ref{fig:multi-object_interation}.

\subsection{Ablation Study of Video Editing}

\begin{figure*}[h]
  \centering
  \includegraphics[width=\linewidth]{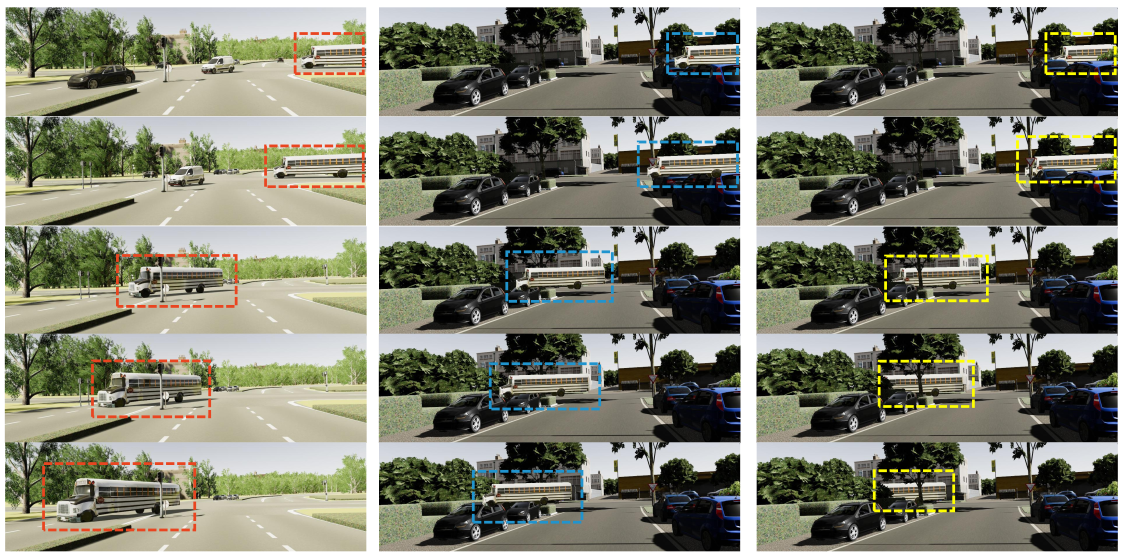}
  \caption{\textbf{Ablation study of Video Editing.} Copy the moving objects from the original video to the current moving video. From left to right: the original video, the edited results of the baseline method, and the edited results of the PVO method. PVO can model the complete scene information, such as depth, pose, optical flow, panoptic segmentation, etc. Compared to the baseline method, which edits the scene only by segmentation and optical flow, the results of PVO are more realistic.}
  \label{fig:online_fusion}
\end{figure*}

\begin{figure}[t]
  \centering
  \includegraphics[width=\linewidth]{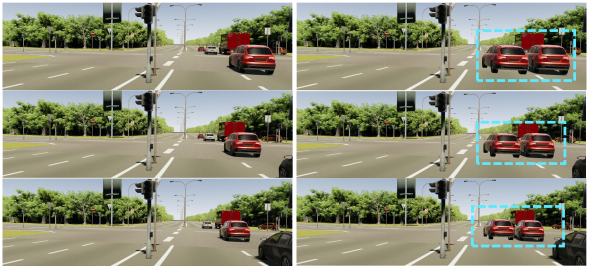}
  \caption{\textbf{Copy \& Paste.} We can replicate the same moving vehicle in a new lane and keep the video consistent, leveraging the proposed PVO method. }
  \label{fig:copy_paste}
\end{figure}

\begin{figure}[t]
  \centering
  \includegraphics[width=\linewidth]{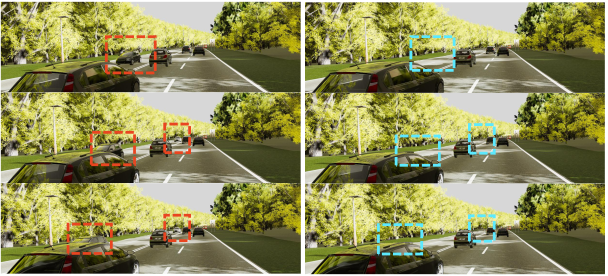}
  \caption{\textbf{Delete.} If a single lane is overloaded, we can also perform the operation of removing the moving vehicle using the proposed PVO method. From left to right: the original video, the edited result of removing the car.}
  \label{fig:Delete}
\end{figure}
We perform an ablation study of PVO in video editing compared with the existing method.

Baseline: We use Video Propagation Network \cite{jampani2017video} to perform video editing for motion control. The baseline method is generally simple to manipulate objects without considering occlusion, but it doesn't look realistic. 

Ours (PVO): The PVO method can better model scene segmentation and motion geometry information and achieve better object manipulation results in occluded scenes, shown in Fig.~\ref{fig:online_fusion}.

\subsection{Motion Control}
As shown in Fig.~\ref{fig:motion_controler}, we 
can insert moving objects into the new scene and also directly manipulate the motion patterns of the moving objects of the original video, such as acceleration, deceleration, pause, and rewind. We can also apply PVO to natural scenes such as Cityscapes for motion control, shown in Fig.~\ref{fig:general motion} which shows the generalization of Panoptic Visual Odometry.


\subsection{Single vs Multi Instance Interaction} 
Since PVO provides more comprehensive information about the motion and panoptic segmentation of the scene for the cases such as occlusion, where adjacent frames can be complemented, we can better perform the interaction between different instances, as shown in Fig.~\ref{fig:multi-object_interation}.

\subsection{Copy and Paste}
It's also interesting to copy objects running in other lanes into an empty lane, shown in Fig.~\ref{fig:copy_paste}.

\subsection{Delete} 
If a single lane is overloaded, we can also perform the operation of removing the running vehicle, shown in Fig.~\ref{fig:Delete}.

\section{Discussion}
\label{sec:discussion}


Although PVO can model the panoptic segmentation and motion information of the scene well and support video editing effects such as manipulating the motion patterns of objects. However, it does not take into account the intrinsic physical information of the scene~\cite{Ye2022IntrinsicNeRF}, such as lighting, materials, shading, etc., so it cannot make a completely realistic video. In addition, effects~\cite{lu2021omnimatte} related to the objects themselves, such as shadows, cannot be modeled. Fully modeling the movement, panoptic segmentation, effects, and physical properties of the scene is an issue worth exploring. Furthermore, we can explore the application of PVO to autonomous driving simulations to test the robustness of autonomous driving systems by manipulating the motion of objects. We leave this as future work.

\end{document}